\documentclass[10pt,twocolumn,letterpaper]{article}

\usepackage[pagenumbers]{cvpr} 

\definecolor{cvprblue}{rgb}{0.21,0.49,0.74}
\usepackage[pagebackref,breaklinks,colorlinks,allcolors=cvprblue]{hyperref}
\newcommand{\benchmark}{{\sc SemSegBench}}
\usepackage{booktabs}
\usepackage{algpseudocode}
\usepackage{algorithm}
\usepackage{multirow}
\usepackage{multicol}
\usepackage{mathtools}
\usepackage{ragged2e}
\usepackage[export]{adjustbox}
\usepackage{setspace}

\newcommand\blfootnote[1]{%
  \begingroup
  \renewcommand\thefootnote{}\footnote{#1}%
  \addtocounter{footnote}{-1}%
  \endgroup
}

\usepackage[capitalize]{cleveref}
\crefname{section}{Sec.}{Secs.}
\Crefname{section}{Section}{Sections}
\Crefname{table}{Table}{Tables}
\crefname{table}{Tab.}{Tabs.}
\definecolor{cadmiumgreen}{rgb}{0.0, 0.42, 0.24}
\definecolor{custom}{cmyk}{0.1,0.48,0.49,0.2}
\definecolor{OliveGreen}{cmyk}{0.64,0,0.95,0.40}
\definecolor{new}{rgb}{0.81,0.05,0.9}
\definecolor{BrickRed}{rgb}{0.81,0.1,0.1}
\definecolor{RoyalBlue}{rgb}{0.2,0.2,0.75}

   \usepackage[normalem]{ulem}
   \usepackage{bm}

\makeatletter
\DeclareRobustCommand\onedot{\futurelet\@let@token\@onedot}
\def\@onedot{\ifx\@let@token.\else.\null\fi\xspace}

\def\eg{e.g\onedot}

\makeatother

\def\clap#1{\hbox to 0pt{\hss #1\hss}}%
\def\initials#1{\protect\clap{\protect\smash{\protect\raisebox{1.4ex}{\protect\tiny{\protect\textsf{\protect\textit{#1}}}}}}}%
\makeatletter
\newcommand{\EDIT}[4][]{\protect\@ifundefined{hidecomments}{%
  \protect\strut{\color{#3}{\hspace{0pt}\initials{#2}\protect\sout{#1}{~#4}}}%
  }{#4}}
\newcommand{\NOTEboxed}[3]{\protect\@ifundefined{hidecomments}{%
  {\begin{center}\fbox{\parbox{0.97\linewidth}{\protect\EDIT{#1}{#2}{#3}}}\end{center}}
  }{}}
\newcommand{\COMM}[3]{\protect\@ifundefined{hidecomments}{%
  {\protect\EDIT{#1}{#2}{#3}}%
  }{}}
\newcommand{\DefAuthor}[2] 
{%
  \expandafter\newcommand\csname #1edit\endcsname[2][]{\protect\EDIT[##1]{#1}{#2}{##2}}
  \expandafter\newcommand\csname #1\endcsname[1]{\protect\COMM{#1}{#2}{[##1]}}
  \expandafter\newcommand\csname #1boxed\endcsname[1]{\protect\NOTEboxed{#1}{#2}{##1}}
}
\definecolor{dfltgreen}       {rgb}{0.0,0.5,0.0}
\definecolor{dfltred}         {rgb}{0.7,0.0,0.0}
\newcommand{\REVadd}[1]{\protect\@ifundefined{hidecomments}{%
  \strut{\color{dfltgreen}{#1}}}{#1}}
\newcommand{\REVedit}[2][]{\protect\@ifundefined{hidecomments}{%
  \strut{\color{dfltred}{\protect\sout{#1}}\color{dfltgreen}{~#2}}}%
  {#2}}
\makeatother

\definecolor{dkgreen}       {rgb}{0.0,0.5,0.0}
\definecolor{dkblue}        {rgb}{0.0,0.0,0.7}
\definecolor{dkcyan}        {rgb}{0.0,0.5,0.5}
\definecolor{dkmagenta}     {rgb}{0.5,0.0,0.5}
\DefAuthor{MK}{dkmagenta} 
\DefAuthor{AB}{dkgreen} 
\DefAuthor{JS}{dkblue} 
\DefAuthor{SC}{dkcyan} 
\DefAuthor{SA}{orange} 
\DefAuthor{todo}{red}

\usepackage{pifont}
\def\paperID{0005} 
\def\confName{CVPR}
\def\confYear{2025}

\title{Are Synthetic Corruptions A Reliable Proxy For Real-World Corruptions?}

\author{Shashank Agnihotri$^{*,1}$
\and
David Schader$^{*,1}$
\and
Nico Sharei$^{*,1}$
\and
Mehmet Ege Kaçar$^{*,1}$
\and
Margret Keuper$^{1,2}$\\
$^{1}$Data and Web Science Group, University of Mannheim, Germany \\
$^{2}$Max-Planck-Institute for Informatics, Saarland Informatics Campus, Germany \\
{\tt\small shashank.agnihotri@uni-mannheim.de}
}

\begin{document}

%
\twocolumn[{%
\renewcommand\twocolumn[1][]{#1}%
\maketitle
\begin{center}
\vspace{-.5cm}
    \centering
    \scalebox{0.9}{
    \begin{tabular}{@{}c@{\hspace{1.5mm}}c@{\hspace{1.5mm}}c@{}cc}
    & &
    Input Image & \hfil \phantom{aaaaaaaaaaa} Ground Truth & \hfil Prediction \\
    \multirow{2}{*}[3em]{\rotatebox{90}{Real World Corruption}} &
    \rotatebox{90}{\phantom{aaaaaa} Snow} &
    \includegraphics[width=0.3\linewidth]{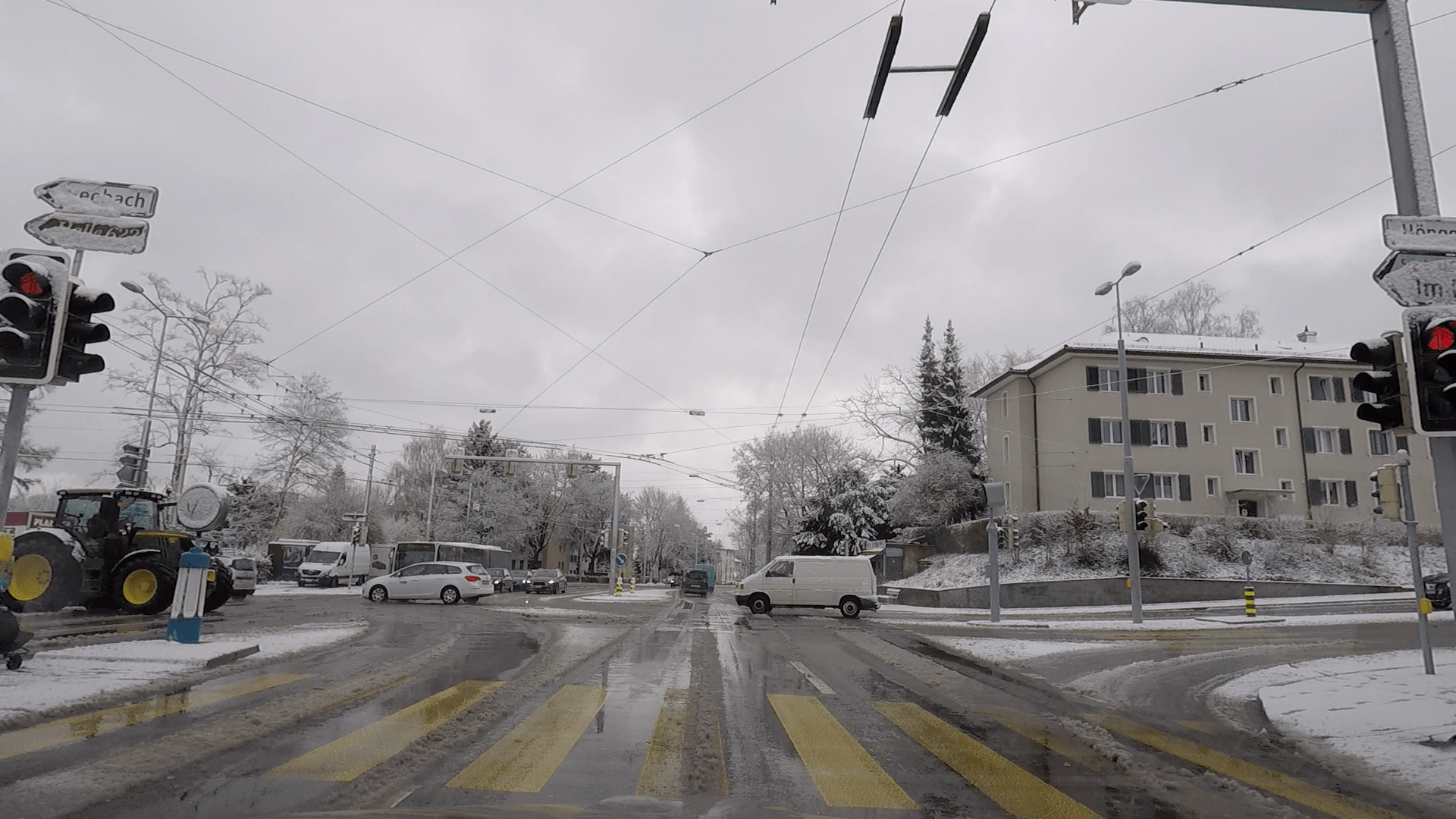}     
         & 
    \multicolumn{2}{c}{\includegraphics[width=0.6\linewidth]{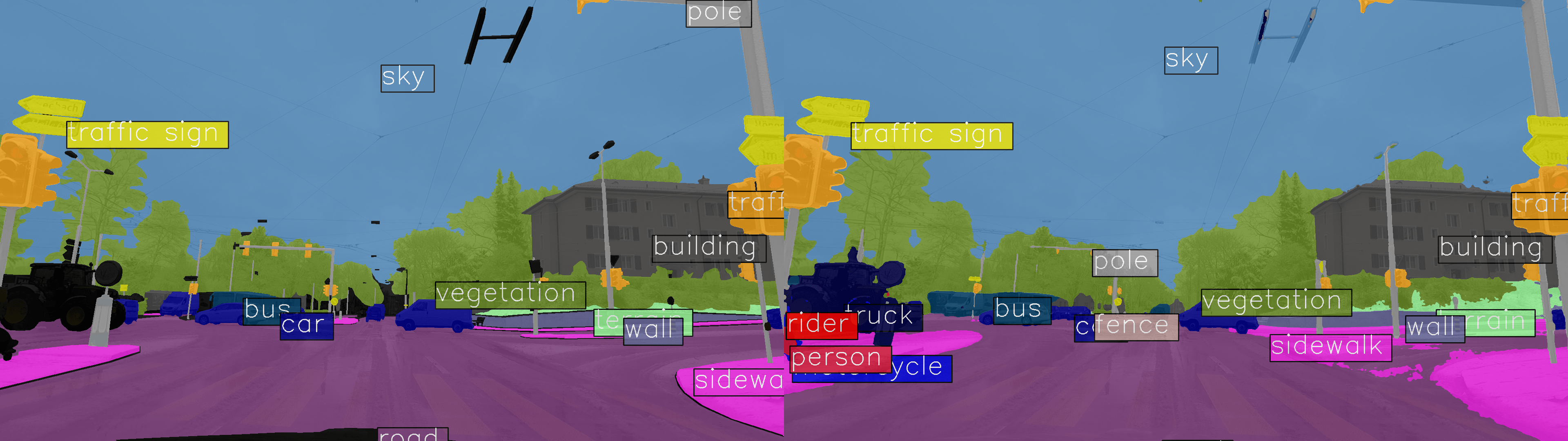}}
         \\

        &
        \rotatebox{90}{\phantom{aaaaaaa} Fog} &
    \includegraphics[width=0.3\linewidth]{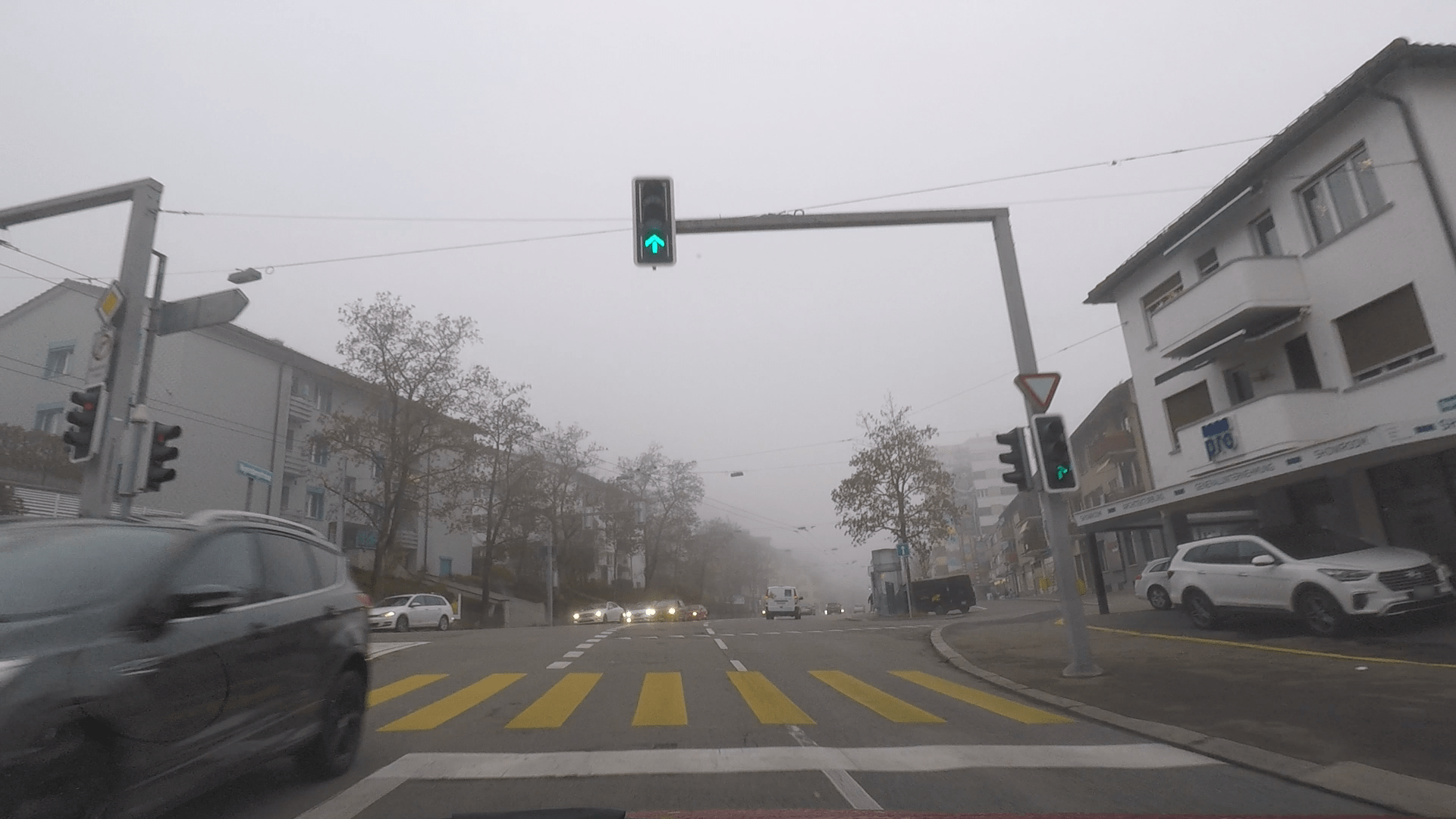}     
         & 
    \multicolumn{2}{c}{\includegraphics[width=0.6\linewidth]{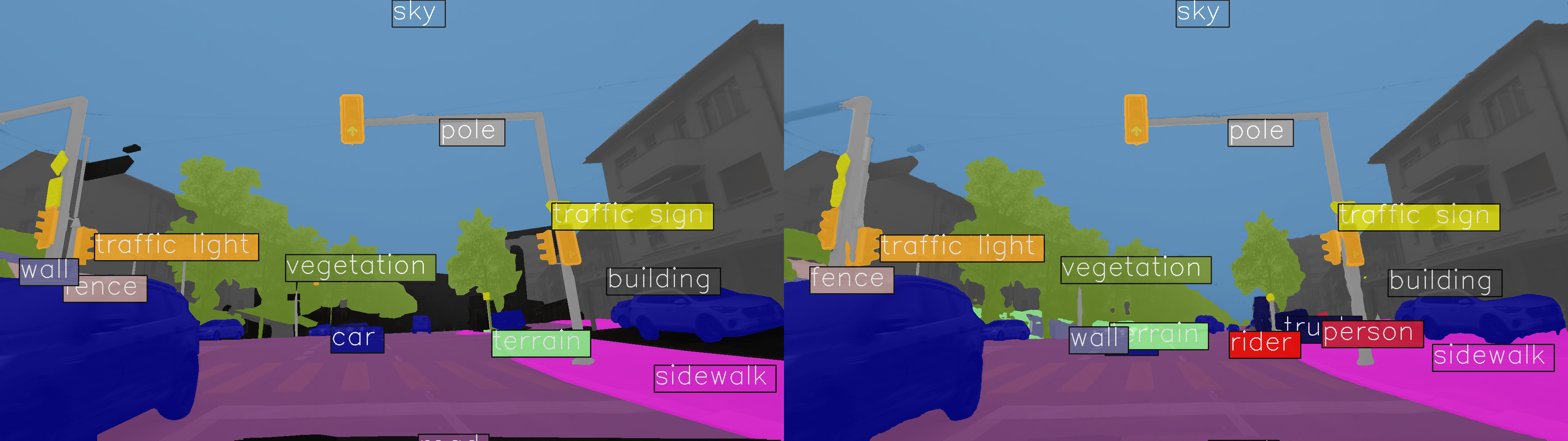}}
         \\


    \multirow{2}{*}[3em]{\rotatebox{90}{Synthetic Corruption}} &
    \rotatebox{90}{\phantom{aaaaa} Snow} &
    \includegraphics[width=0.3\linewidth]{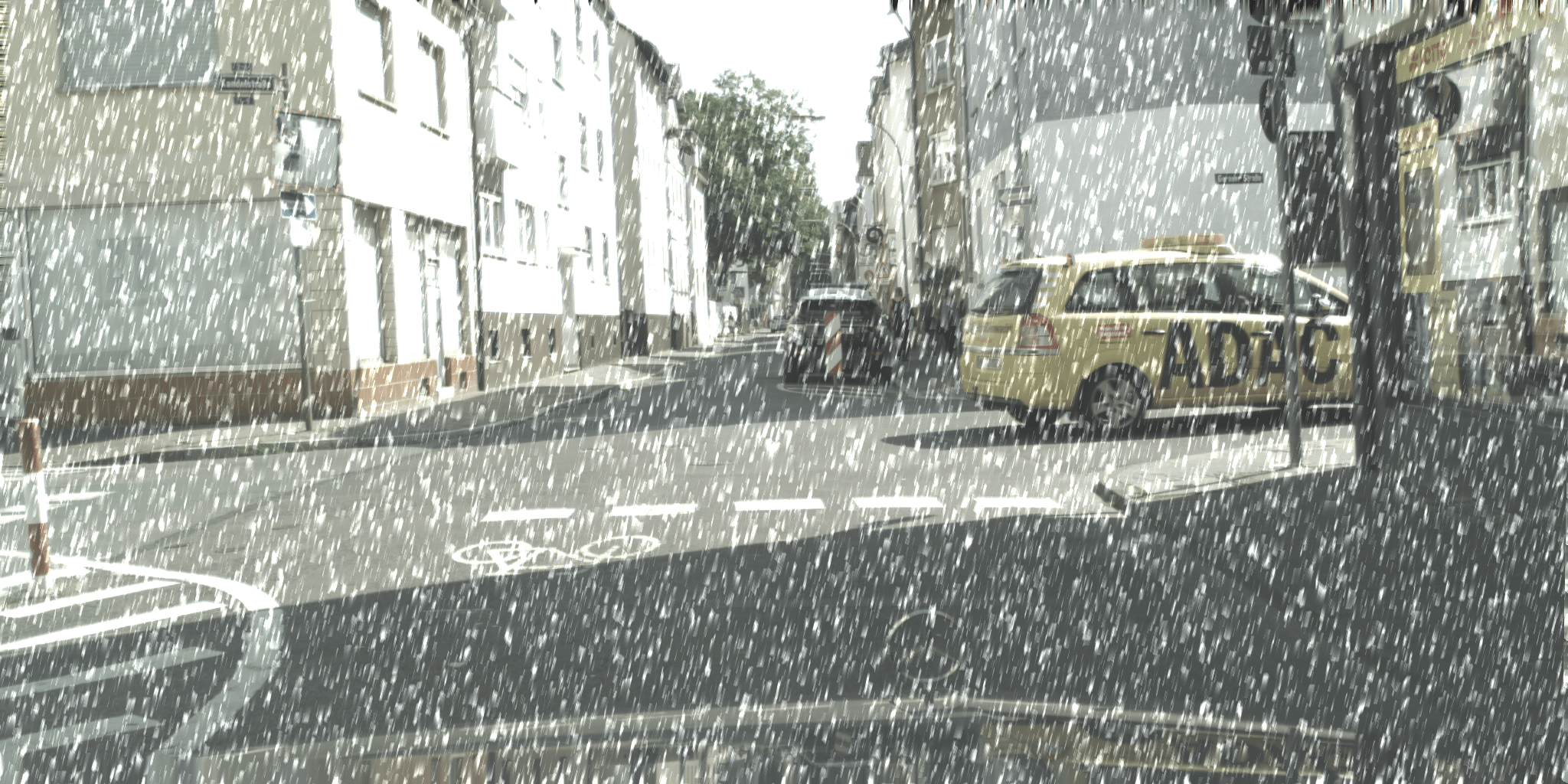}     
         & 
    \multicolumn{2}{c}{\includegraphics[width=0.6\linewidth]{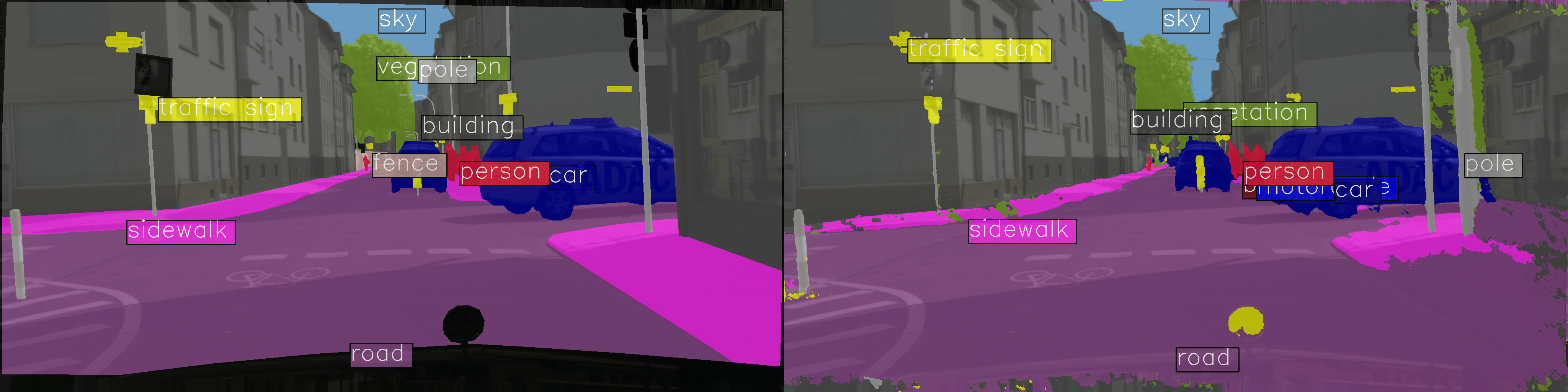}}
         \\

        &
        \rotatebox{90}{\phantom{aaaaaa} Fog} &
    \includegraphics[width=0.3\linewidth]{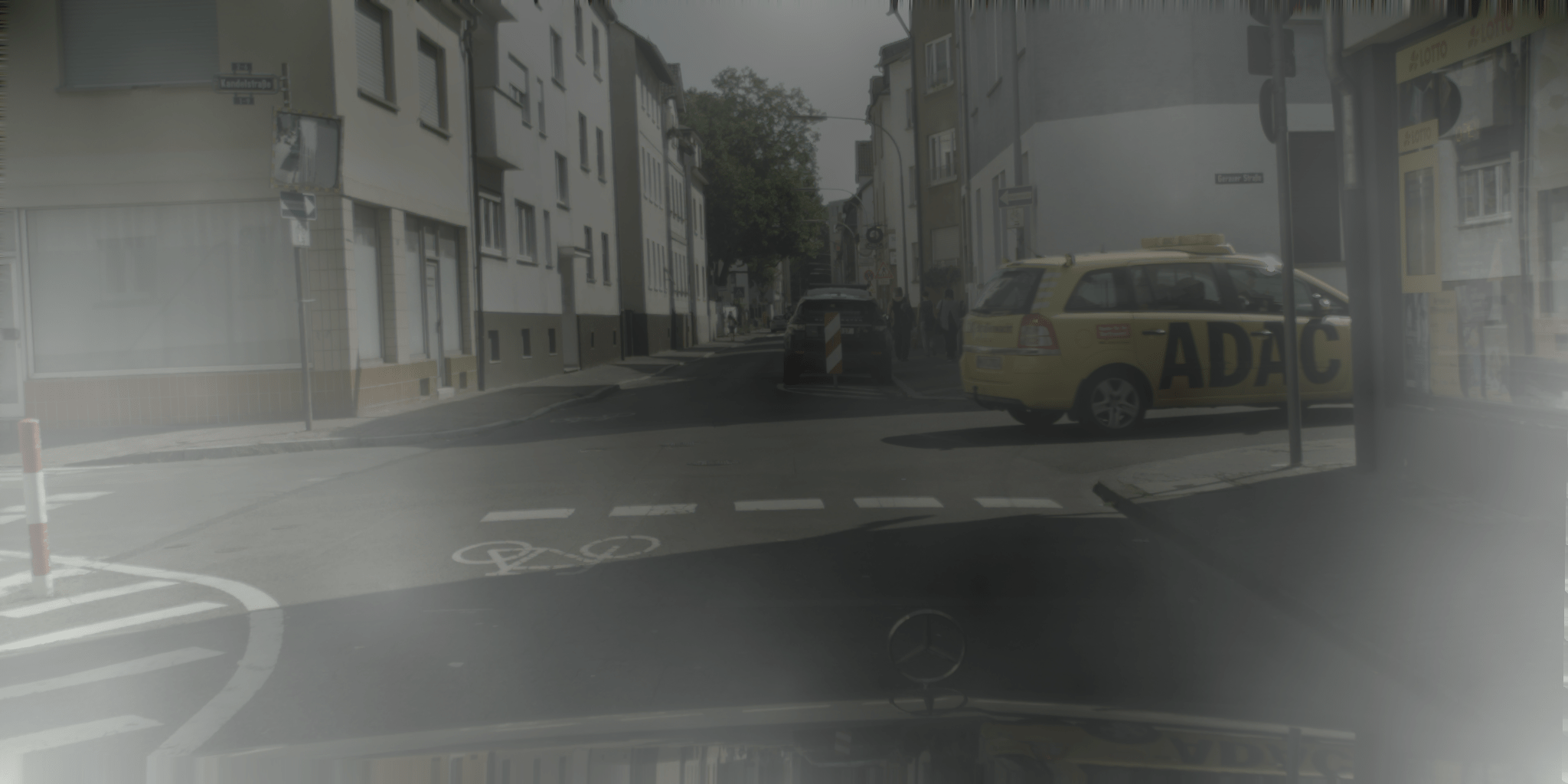}     
         & 
    \multicolumn{2}{c}{\includegraphics[width=0.6\linewidth]{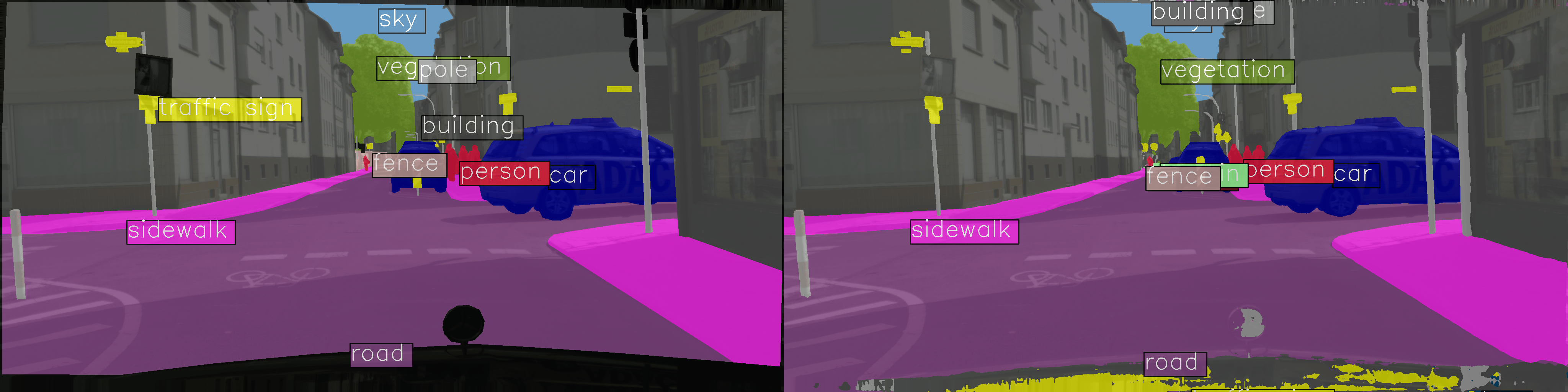}}
         \\
    \end{tabular}
    
    }
    \captionof{figure}{Comparing images with weather corruptions captured in the wild (ACDC~\cite{acdc}) and images corrupted using synthetic corruptions~\cite{commoncorruptions} and the predictions using a Mask2Former~\cite{cheng2021mask2former} with a Swin-Base~\cite{liu2021Swin} backbone trained on the Cityscapes~\cite{cordts2016cityscapes} dataset.}
    \label{fig:teaser}
\end{center}
}]
\begin{abstract}
Deep learning (DL) models are widely used in real-world applications but remain vulnerable to distribution shifts, especially due to weather and lighting changes. Collecting diverse real-world data for testing the robustness of DL models is resource-intensive, making synthetic corruptions an attractive alternative for robustness testing. However, are synthetic corruptions a reliable proxy for real-world corruptions? To answer this, we conduct the largest benchmarking study on semantic segmentation models, comparing performance on real-world corruptions and synthetic corruptions datasets. Our results reveal a strong correlation in mean performance, supporting the use of synthetic corruptions for robustness evaluation. We further analyze corruption-specific correlations, providing key insights to understand when synthetic corruptions succeed in representing real-world corruptions. \href{https://github.com/shashankskagnihotri/benchmarking_robustness/tree/segmentation_david/semantic_segmentation}{Open-source Code}.
\end{abstract}
\blfootnote{Accepted at CVPR 2025 SynData4CV Workshop. $^{*}$Equal Contribution. M.K.~acknowledges funding by the DFG Research Unit 5336. We acknowledge support by the state of Baden-Württemberg through bwHPC.}

\section{Introduction}
\label{sec:intro}


Although very successful in benchmark scenarios, the reliability of deep-learning (DL)-based models for semantic segmentation in real-world scenarios remains a major concern~\cite{das20212,das2023weakly,agnihotri2023cospgd,li2023intra,li2024adversarial}. 
Potentially unseen variations in the data (a.k.a.~distribution shifts), for example, due to changes in weather conditions (\eg, fog, rain, snow) and lighting (\eg, nighttime, glare), can heavily degrade model performance. 
Ensuring robustness to such shifts is critical for safe and reliable deployment, particularly in 
applications like autonomous driving~\cite{kitti15,cordts2016cityscapes,agnihotri2023improving} or medical imaging~\cite{unet,duck_medical,keuper2011hierarchical}. 
To evaluate model robustness, researchers often rely on synthetic corruptions, such as~\cite{commoncorruptions}. 
These perturbations — designed to mimic real-world conditions — offer a scalable and controlled way to assess model performance without the cost of real-world data collection. 

Several previous works \cite{acdc,SegmentMeIfYouCan,sommerhoff2023differentiable,agnihotri2024beware,kamann2020benchmarking_semseg_ood,grabinski2022frequencylowcut,grabinski2022aliasing,grabinski2022robust,agnihotri_dispbench,agnihotri2023unreasonable,agnihotri2024roll,sommerhoff2024task,hoffmann2021towards} have also attempted to draw focus towards threats posed in real-world applications when facing slight domain shifts, for example, through noise or simply through changing weather. 
Specific evaluations involve the study of Out-Of-Distribution (OOD) samples to mimic realistic domain shifts.

Despite their widespread use~\cite{li2024intra}, the correlation between model performance on synthetic and real-world corruptions is not well understood.
\Cref{fig:teaser} shows one such scenario with real-world corruptions (Snow and Fog) captured in the ACDC dataset~\cite{acdc} and similar synthetic corruptions added on in-domain images from the cityscapes validation dataset.
We observe very similar trends in the lack of robustness of the model towards both real-world and synthetic corruptions.
However, a fundamental question remains:
\begin{quote}
    ``Are synthetic corruptions a reliable proxy for real-world corruptions?''
\end{quote}
 
If a strong correlation exists, synthetic corruptions could serve as a cost-effective alternative for robustness evaluation. 
Conversely, if the correlation is weak, extensive tests on real-world settings remain necessary at all stages.

Here, we conduct a large benchmarking study, analyzing the correlation between model performance on real-world and synthetic corruptions for semantic segmentation.
The main contributions of this work are as follows:
\begin{itemize}
    \item We benchmark multiple DL-based semantic segmentation models on real-world corruptions from the ACDC dataset and synthetic corruptions from Cityscapes + 2D Common Corruptions.
   
    
    \item We provide an in-depth analysis of corruption-specific trends, identifying cases where synthetic corruptions succeed or fail as proxies.
    
   \item We provide benchmarking of semantic segmentation methods against synthetic corruptions on ADE20k~\cite{ade20k} and PASCAL VOC 2012~\cite{pascal-voc-2012} datasets.
    
\end{itemize}

Our findings reveal a high correlation in mean performance, suggesting that synthetic corruptions can indeed serve as a reliable proxy for real-world robustness evaluation. 
However, we also highlight key cases where synthetic corruptions fail to fully capture real-world effects, underscoring the need for more nuanced evaluation methods.

\section{Related Work}
\label{sec:related}
The robustness of DL-based methods to distribution shifts is often used as a measure of their generalization ability~\cite{hendrycks2020augmix,hoffmann2021towards,yue2024improving,das20212}. Common Corruptions~\cite{commoncorruptions} and 3D Common Corruptions~\cite{3dcommoncorruptions} are tools proposed for benchmarking the robustness of image classification models~\cite{medi2024towards,medi2025fair,prasse2024aligning}, but they can be extended to other vision tasks as for example done in \cite{kamann2020benchmarking_semseg_ood}.
However, both are synthetic corruptions, and distribution shifts occurring in the real world might be slightly different. 
Conversely, \citet{acdc} proposed ``ACDC: The Adverse Conditions Dataset with Correspondences for Robust Semantic Driving Scene Perception''.
This dataset contains images captured in the wild in different conditions, such as during Night, Rain, Snow, and Fog.
While ACDC does not cover many other possible conditions that can cause distribution shifts, it serves as a community-accepted tool for benchmarking real-world OOD robustness to a certain extent.

In this work, we use both Common Corruptions and ACDC to benchmark OOD robustness and thus measure the generalization ability of various semantic segmentation methods, including recently proposed SotA methods like Mask2Former~\cite{cheng2021mask2former} and InternImage~\cite{wang2023internimage}, with the goal to investigate whether synthetic datasets that are easy to generate can serve as a proxy for a model's real world OOD robustness.

\cite{SegmentMeIfYouCan} provides a new benchmark for robustness against anomalies, while relevant for real-world applications, we intend to focus this work on traditional OOD robustness.

In their work, \citet{cityscapes-c} proposed datasets combining 2D Common Corruptions with datasets such as MS-COCO~\cite{ms-coco}, PASCAL VOC 2007~\cite{pascal_voc_2007}, and Cityscapes. 
However, their evaluations were limited to 2D Common Corruptions and how different severities of the corruptions on the images impact the downstream task performance.
We find correlations between performance against 2D Common Corruptions and real-world corruptions.
We use their proposed Cityscapes-C (Cityscapes + 2D Common Corruptions) as our synthetic corruptions dataset. 



\section{Metrics For Analysis At Scale}
\label{sec:metrics}
This is the first work to analyze semantic segmentation methods, especially under the lens of reliability and generalization ability on such a large scale.
The most commonly used metrics for reporting evaluations on semantic segmentation are mean Intersection over Union (mIoU), mean class Accuracy (mAcc), and mean Accuracy of all pixels (aAcc)~\cite{semsegzhao2017pspnet,semseg_adv,agnihotri2023cospgd,schneider2024implicit}.
We capture these metrics while evaluating models against both ACDC and the 15 2D Common Corruptions on the Cityscapes validation dataset.
As per the commonly accepted practice of such OOD evaluations, all models are pre-trained on the Cityscapes training dataset.

Similar to \cite{cityscapes-c}, the 15 2D Common Corruptions~\cite{commoncorruptions} considered in this work are: `gaussian noise’, `shot noise’, `impulse noise’, `defocus blur’, `frosted glass blur’, `motion blur’, `zoom blur’, `snow’, `frost’, `fog’, `brightness’, `contrast’, `elastic’, `pixelate’, `jpeg’.
Similar to \cite{commoncorruptions}, \citet{cityscapes-c} shows that synthetic corruptions with corruption severity=1 are too weak, and corruptions with corruption severity=5 are too strong for the downstream task.
Thus, we use corruption severity=3 in our evaluations.

As discussed, multiple image classification works~\citep{robustbench,hendrycks2020augmix,hoffmann2021towards} and some semantic segmentation works~\cite{kamann2020benchmarking_semseg_ood,cityscapes-c} use OOD Robustness of models for evaluating the generalization ability of the method.
However, different image corruptions impact the performance of the semantic segmentation methods differently.
As we are interested in the worst possible case, we define $\mathrm{Generalization\text{ }Ability\text{ }Measure}$ ($\mathrm{GAM}$) as the worst mIoU across all image corruptions at a given severity level.
That is, we ask the question ``For a given dataset, what is the worst possible performance of a given method?''.
Answering this question tells us about the reliability and generalization ability of a method.
We find the minimum of the mIoU of the segmentation masks predicted under image corruptions w.r.t. the ground truth masks for a given method, across all corruptions at a given severity and report this as the $\mathrm{GAM}_{\mathit{severity\ level}}$.
For example, for severity=3, the measure would be denoted by $\mathrm{GAM}_{3}$.
The higher the $\mathrm{GAM}$ value, the better the generalization ability of the given semantic segmentation method.
In \cref{sec:appendix:correlation_metrics}, we show that our observations are not limited to the mIoU metric and extend to other metrics as well.



\section{Analysis And Key Findings}
\label{sec:analysis}
We analyze the correlation in mean performance to determine whether synthetic corruptions can serve as a reliable proxy for real-world corruptions. Additionally, we conduct an in-depth examination of corruption-specific trends, identifying cases where synthetic corruptions effectively mimic real-world effects and where they fall short.

\subsection{Are Synthetic Corruptions Useful?}
\label{subsec:discussion:2dcc_acdc}
\begin{figure*}
    \centering
    \includegraphics[width=0.9\linewidth, trim={0 0.5cm 0 0},clip]{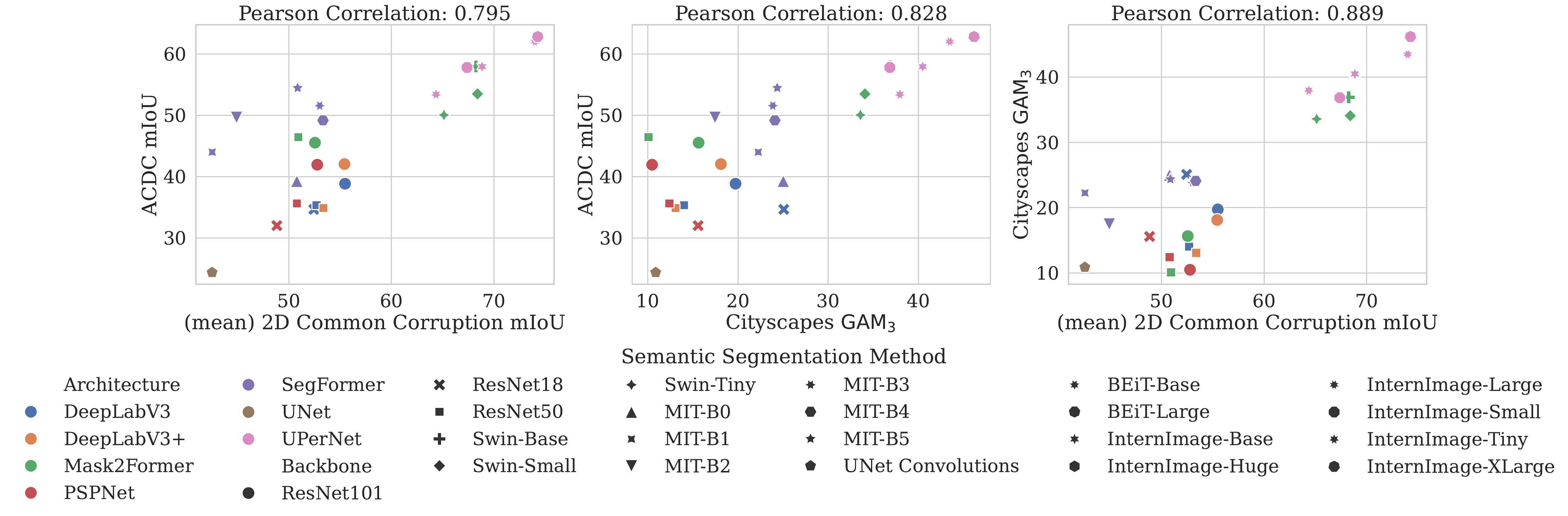}
    \caption{To empirically determine if synthetic common corruptions such as those proposed by \cite{commoncorruptions} truly represent the distribution and domain shifts in the real world, we try to find correlations in evaluations on ACDC and 2D Common Corruptions. Each model is trained on the training dataset of the Cityscapes dataset. 
    Left plot: The y-axis represents values from evaluations on the ACDC dataset, and the x-axis represents mean performance from evaluations on the Common Corruptions at severity=3. We observe a high positive correlation.
    Centre plot: The y-axis again represents values from evaluations on the ACDC dataset, while the x-axis represents $\mathrm{GAM}_3$, which is the worst performance of the methods across all the Common Corruptions at severity=3. We observe a slightly higher positive correlation.
    Right plot: serves as a sanity check, where the y-axis represents $\mathrm{GAM}_3$ and the x-axis represents mean performance from evaluations on the Common Corruptions at the same severity.
    We observe a very high correlation in performance.
    Thus, given the high positive correlations between performance on the ACDC and mean performance against all synthetic common corruption, we conclude for relative analysis that synthetic corruptions do serve as a reliable proxy for real-world corruptions.}
    \label{fig:correlation_2dcc_acdc}
\end{figure*}
We attempt to study if synthetic corruption like that introduced by \cite{commoncorruptions} does represent the distribution shifts in the real world.
While this assumption has driven works such as \cite{commoncorruptions,3dcommoncorruptions,kamann2020benchmarking_semseg_ood}, to the best of our knowledge, it has not yet been proven.
Previous works on robustness~\cite{guo2023robustifying} simply report performance on both, thus, to save compute in the future, we prove this assumption in \cref{fig:correlation_2dcc_acdc}.

For this analysis, we used methods trained on the training set of Cityscapes and evaluated them on 2D Common Corruptions~\cite{commoncorruptions} and the ACDC datasets.
ACDC is the Adverse Conditions Dataset with Correspondences, consisting of images from similar regions and scenes as Cityscapes but captured under different conditions such as Day/Night, Fog, Rain, and Snow. 
These are corruptions in the real world, thus, we attempt to find correlations between performance against synthetic corruptions from 2D Common Corruptions (severity=3) and ACDC.
We analyze each common corruption separately and also the mean performance across all 2D Common Corruptions.

In \cref{fig:correlation_2dcc_acdc}, we observe a very strong positive correlation in performance against ACDC and mean performance across all 2D Common Corruptions.
This novel finding helps the community significantly. 
It means that we do not need to go into the wild to capture images with distribution shifts, as synthetic corruptions serve as a reliable proxy for real-world conditions.
Next, we look at the correlation between the worst-case scenario measure using $\mathrm{GAM}_3$ and ACDC.
Here, we observe a higher correlation than the previous case, indicating that the performance against worst-case corruption serves as a reliable proxy for real-world corruptions.
Lastly, as a sanity check, we find the correlation between mean performance against all corruptions and performance against worse-case corruption to observe a very high correlation.
Showing that the two can be used interchangeably. 

\subsection{When Do Synthetic Corruptions Succeed?}
\label{subsec:analysis:success_failure}
\begin{figure*}
    \centering
    \includegraphics[width=0.9\linewidth]{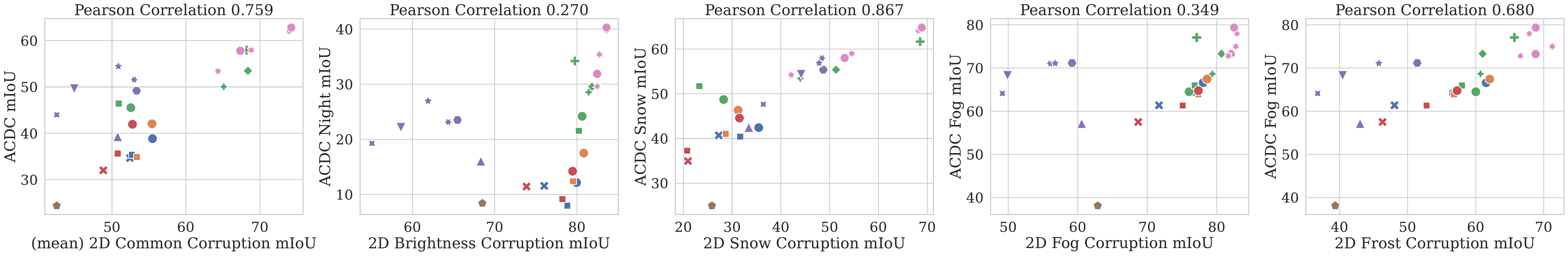}
    \caption{Correlation between model performance (legend as in \cref{fig:correlation_2dcc_acdc}) on ACDC (real-world corruptions) and 2D Common Corruptions (synthetic) for different corruption types. The left-most plot shows the correlation between mean mIoU across all 2D Common Corruptions and ACDC, with a strong Pearson correlation of 0.759, indicating that synthetic corruptions are generally a reasonable proxy for real-world robustness. The remaining plots analyze specific corruptions: brightness (synthetic) vs. night (real) with correlation 0.270, snow (synthetic) vs. snow (real) with correlation 0.867, fog (synthetic) vs. fog (real) with correlation 0.349, and frost (synthetic) vs. fog (real) with correlation 0.680. While some synthetic corruptions (e.g., snow) closely align with their real-world counterparts, others (e.g., brightness for night) exhibit weaker correlations, highlighting cases where synthetic corruptions may fail as accurate proxies.}
    \label{fig:correlation_2dcc_acdc_success_failure}
\end{figure*}
Since some synthetic corruptions attempt to directly mimic the real-world scenarios in ACDC, like changes in lighting due to Day/Night changes or changes in weather due to snowfall or fog, we analyze the correlation of relevant corruptions to ACDC.
As discussed in \cref{subsec:discussion:2dcc_acdc}, the mean performance correlation is high.
However, we observe in \cref{fig:correlation_2dcc_acdc_success_failure} that individual corruptions exhibit varying levels of agreement between synthetic and real-world effects.
We observe that the Snow corruption shows a very strong alignment (Pearson correlation 0.867), indicating that synthetic snow corruptions effectively mimic real-world snow-related degradation, despite the corrupted images looking different to a human observer (as shown in \cref{fig:teaser}).

Brightness (Pearson correlation 0.270) and Fog (Pearson correlation 0.349) exhibit weak alignment, suggesting that synthetic versions of these corruptions fail to fully capture real-world complexities.
Specifically, brightness corruptions struggle to model real-world nighttime conditions, while synthetic fog does not accurately represent atmospheric distortions seen in real-world data.

These findings highlight that while synthetic corruptions can approximate real-world robustness trends, they are not universally reliable across all corruption types.

Interestingly, we observe a moderate positive correlation (Pearson correlation 0.680)  in performance against ACDC Fog and 2D Common Corruption Frost.
Since the Frost 2D Common Corruption involves superimposing a randomly chosen frost image on the input image with some transparency, one might hypothesize that the model finds the distribution shifts between the two to be moderately similar.

\section{Conclusion}
\label{sec:conclusion}
Our study provides the most comprehensive benchmarking to date on the reliability of synthetic corruptions as a proxy for real-world distribution shifts in semantic segmentation. Through extensive experiments, we observe a strong correlation in mean performance between synthetic and real-world corruptions, supporting their utility for robustness evaluation. However, a deeper analysis of individual corruption types reveals that while some synthetic corruptions (e.g., snow) closely align with real-world performance, others (e.g., brightness, fog) exhibit weak correlations, highlighting gaps in current benchmarking approaches.

These findings underscore the importance of refining synthetic corruption benchmarks to better capture real-world conditions.  
To promote OOD evaluations on synthetic datasets, we provide benchmarking of all 15 2D Common Corruptions on the most commonly used semantic segmentation datasets, namely, Cityscapes, ADE20k, and PASCAL VOC2012 datasets.

{
    \small
    \bibliographystyle{ieeenat_fullname}
    \bibliography{main}
}
\newpage
\appendix
\onecolumn
{
    \centering
    \Large
    \textbf{Are Synthetic Corruptions A Reliable Proxy For Real-World Corruptions?} \\
    \vspace{0.5em}Paper \#0005 Supplementary Material \\
    \vspace{1.0em}
}

\section*{Table Of Content}
The supplementary material covers the following information:
\begin{itemize}
\setlength\itemsep{2em}
    \vspace{1em}
    \item \Cref{sec:appendix:correlation_metrics}: Here we show a high positive correlation in the different metrics captures for correlation between performance against real-world corruptions and synthetic corruptions.
    \item \Cref{sec:appendix:implementation_details}: Additional implementation details for the evaluated benchmarking, such as:

    \begin{itemize}
     \item \Cref{subsec:appendix:details:compute_resources}: Compute resources used.
     
        \item \Cref{sec:appendix:dataset_details}: Details for the datasets used.

        \begin{itemize}
    \setlength\itemsep{1em}
    \vspace{0.5em}
        \item \Cref{subsec:appendix:dataset_details:ade20k}: ADE20K
        \item \Cref{subsec:appendix:dataset_details:cityscapes}: Cityscapes
        \item \Cref{subsec:appendix:dataset_details:pascal_voc}: PASCAL VOC2012        
    \end{itemize}

    \item \Cref{sec:appendix:model_zoo}: A comprehensive look-up table for all the semantic segmentation methods' model weight and datasets pair available in \benchmark{} and used for evaluating the benchmark.
    
    \end{itemize}
    
    \item \Cref{subsubsec:appendix:description:2dcc}: Description of the 2D Common Corruptions used and visualizations of some corruptions on the Cityscapes validation dataset and the performance of InternImage-Base on these corrupted images.

    \item \Cref{sec:appendix:additional_results}: Here we provide benchmarking results from 2D Common Corruption evaluations at severity 3, for the ADE20K, Cityscapes, and PASCAL VOC2012 datasets.

    \item \Cref{sec:related_work_extension}: Extension To Related Work: Here, we extend the related work to discuss a few other important works.

    \item \Cref{subsec:conclusion:future_work} Future Work: Following, we discuss the future directions possible from this work and extension of this work.

    \item \Cref{subsec:conclusion:limitations} Limitations: We discuss the limitations of this work in detail.
\end{itemize}
\vspace{1em}

Due to significant similarity, some of the text here has been adapted from \cite{flowbench}.

\section{Correlation In Metrics}
\label{sec:appendix:correlation_metrics}
\begin{figure}
    \centering
    \begin{tabular}{c}
         \includegraphics[width=1.0\linewidth]{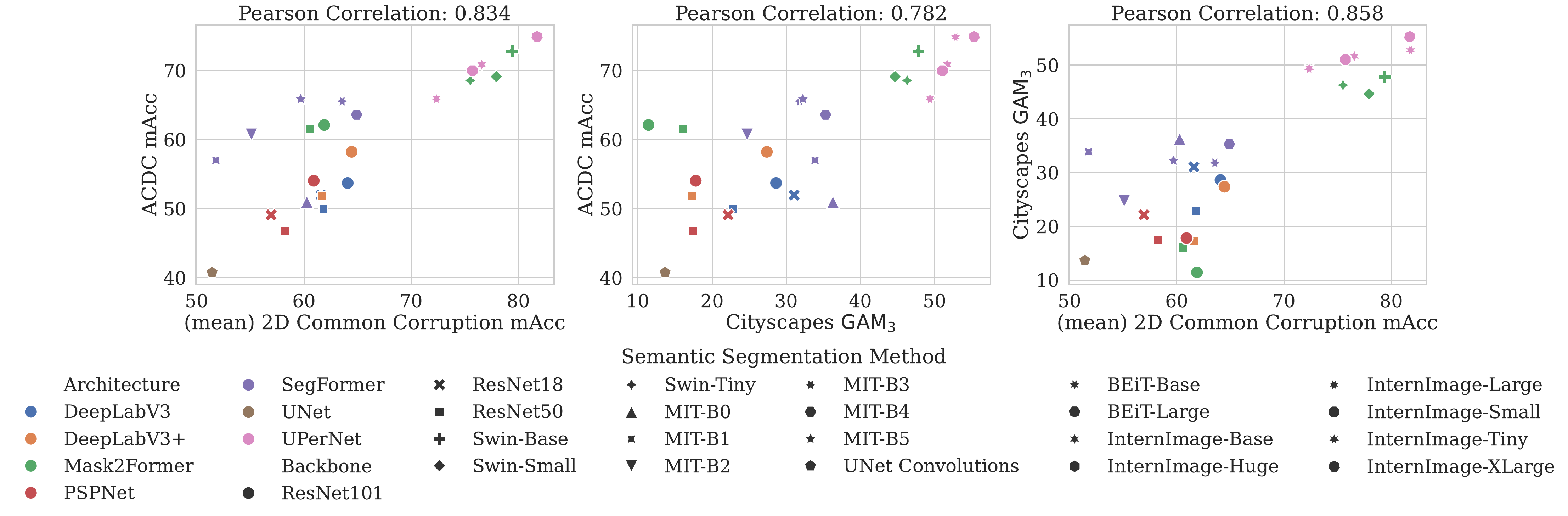}  \\
          \includegraphics[width=1.0\linewidth]{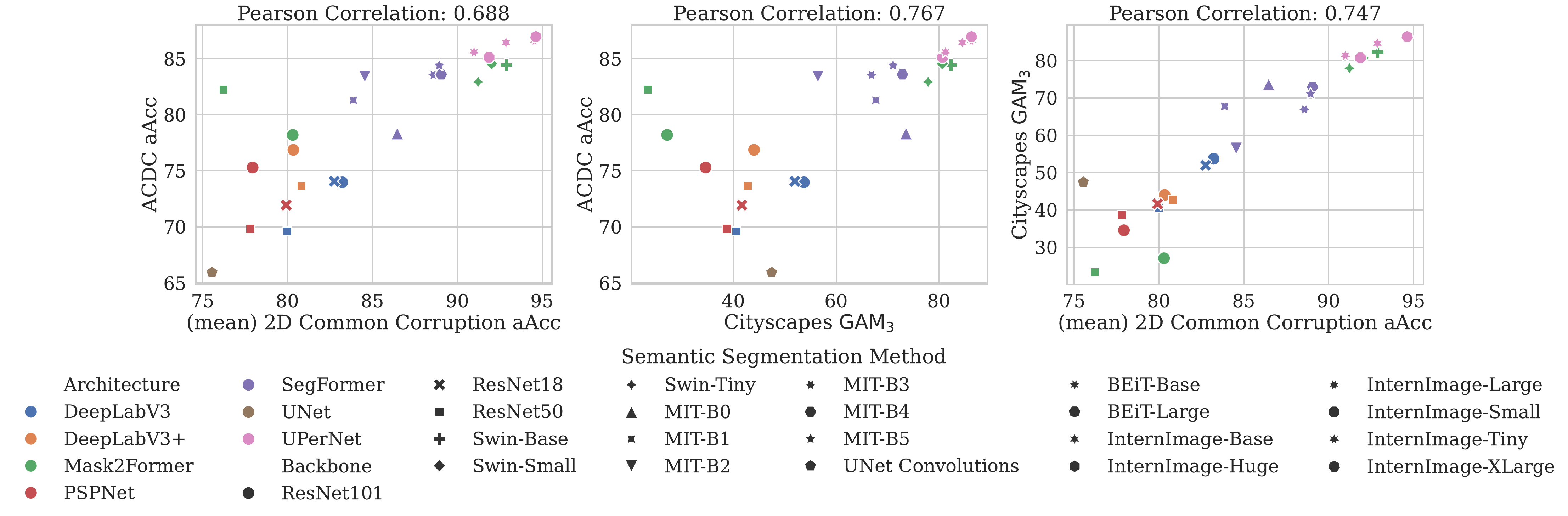}  \\
    \end{tabular}
    \caption{Comparison of mean accuracy across synthetic (2D Common Corruptions) and real-world (ACDC) corruptions. The top plot presents mAcc (mean class accuracy) with a stronger correlation of 0.782–0.858, while the bottom plot shows results for aAcc (all pixel accuracy) with a Pearson correlation of 0.688–0.767. These results indicate that synthetic corruptions serve as a reasonable proxy for real-world robustness, even when measured using metrics other than mIoU}
    \label{fig:correlation_in_metrics}
\end{figure}
Here, we provide a comparison of mean accuracy across synthetic (2D Common Corruptions) and real-world (ACDC) corruptions. The top plot presents mAcc (mean class accuracy) with a stronger correlation of 0.782–0.858, while the bottom plot shows results for aAcc (all pixel accuracy) with a Pearson correlation of 0.688–0.767. 
These results indicate that synthetic corruptions serve as a reasonable proxy for real-world robustness.
Thus, the analysis made using mIoU would also hold if made using other metrics.

\section{Implementation Details Of The Benchmarking}
\label{sec:appendix:implementation_details}
Following, we provide details regarding the experiments done for creating the benchmark used in the analysis.

\subsection{Compute Resources. }
\label{subsec:appendix:details:compute_resources}
Most experiments were done on a single 40 GB NVIDIA Tesla V100 GPU each, however, SegFormer~\cite{xie2021segformer} and Mask2Former~\cite{cheng2021mask2former} with large backbones are more compute-intensive, and thus 80GB NVIDIA A100 GPUs or NVIDIA H100 were used for these models, a single GPU for each experiment.
Training some of the architectures with large backbones required using two to four GPUs in parallel.

\subsection{Dataset Details}
\label{sec:appendix:dataset_details}
Performing OOD robustness evaluations is very expensive and compute-intensive.
Thus, for the benchmark, we only use ADE20k, Cityscapes, and PASCAL VOC2012 as these are the most commonly used datasets for evaluation~\cite{agnihotri2023cospgd,xie2021segformer,cheng2021mask2former,semsegzhao2017pspnet,kamann2020benchmarking_semseg_ood}.

\subsubsection{ADE20K}
\label{subsec:appendix:dataset_details:ade20k}
ADE20K~\cite{ade20k} dataset contains pixel-level annotations for 150 object classes, with a total of 20,210 images for training, 2000 images for validation, and 3000 images for testing. 
Following common practice~\cite{agnihotri2023cospgd,xie2021segformer} we evaluate using the validation images.

\subsubsection{Cityscapes}
\label{subsec:appendix:dataset_details:cityscapes}
The Cityscapes dataset \cite{cordts2016cityscapes} comprises a total of 5000 images sourced from 50 different cities in Germany and neighboring countries. The images were captured at different times of the year and under typical meteorological conditions. Each image was subject to pixel-wise annotations by human experts. The dataset is split into three subsets: training (2975 images), validation (500 images), and testing (1525 images).
This dataset has pixel-level annotations for 30 object classes.

\subsubsection{PASCAL VOC2012}
\label{subsec:appendix:dataset_details:pascal_voc}
The PASCAL VOC 2012~\cite{pascal-voc-2012}, contains 20 object classes and one background class, with 1464 training images, and 1449 validation images. 
We follow common practice \cite{contouring, segpgd, semseg2019, semsegzhao2017pspnet}, and use work by~\citet{SBD_BharathICCV2011}, augmenting the training set to 10,582 images. 
We evaluate using the validation set. 

\noindent\paragraph{Calculating the mIoU. } mIoU is the mean Intersection over Union of the predicted segmentation mask with the ground truth segmentation mask.

\subsection{Models Used}
\label{sec:appendix:model_zoo}
\begin{table}[h]
    \centering
    \caption{An Overview of all the semantic segmentation methods used in the benchmark in this work made using \benchmark{}. Each of the mentioned backbones has been evaluated using each of the architectures and datasets mentioned in the row in this table.}
    \scalebox{0.97}{
    \begin{tabular}{lllc}
         \toprule
         \textbf{Backbone} & \textbf{Architecture} & \textbf{Datasets} & \textbf{Time Proposed (yyyy-mm-dd)} \\
            \midrule

            ResNet101~\cite{resnet} & \shortstack{DeepLabV3~\cite{deeplabv3}, DeepLabV3+~\cite{deeplabv3+},\\ Mask2Former~\cite{cheng2021mask2former}, PSPNet~\cite{semsegzhao2017pspnet}}  & \shortstack{ADE20K, Cityscapes,\\ PASCAL VOC 2012}  & 2017-12-05\\
            \midrule
            ResNet18~\cite{resnet} & \shortstack{DeepLabV3~\cite{deeplabv3}, DeepLabV3+~\cite{deeplabv3+},\\ PSPNet~\cite{semsegzhao2017pspnet}}  & Cityscapes  & 2017-12-05\\
            \midrule
            ResNet50~\cite{resnet} & \shortstack{DeepLabV3~\cite{deeplabv3}, DeepLabV3+~\cite{deeplabv3+},\\ Mask2Former~\cite{cheng2021mask2former}, PSPNet~\cite{semsegzhao2017pspnet}}  & \shortstack{ADE20K, Cityscapes,\\ PASCAL VOC 2012}  & 2017-12-05\\
            \midrule
            Swin-Base~\cite{liu2021Swin} & Mask2Former~\cite{cheng2021mask2former}  & \shortstack{ADE20K, Cityscapes,\\ PASCAL VOC 2012}  & 2022-06-15\\
            \midrule
            Swin-Small~\cite{liu2021Swin} & Mask2Former~\cite{cheng2021mask2former}  & \shortstack{ADE20K, Cityscapes,\\ PASCAL VOC 2012}  & 2022-06-15\\
            \midrule
            Swin-Tiny~\cite{liu2021Swin} & Mask2Former~\cite{cheng2021mask2former}  & \shortstack{ADE20K, Cityscapes,\\ PASCAL VOC 2012}  & 2022-06-15\\
            \midrule
            MIT-B0~\cite{xie2021segformer} & SegFormer~\cite{xie2021segformer}  & \shortstack{ADE20K, Cityscapes,\\ PASCAL VOC 2012}  & 2021-10-28\\
            \midrule
            MIT-B1~\cite{xie2021segformer} & SegFormer~\cite{xie2021segformer}  & \shortstack{ADE20K, Cityscapes,\\ PASCAL VOC 2012}  & 2021-10-28\\
            \midrule
            MIT-B2~\cite{xie2021segformer} & SegFormer~\cite{xie2021segformer}  & \shortstack{ADE20K, Cityscapes,\\ PASCAL VOC 2012}  & 2021-10-28\\
            \midrule
            MIT-B3~\cite{xie2021segformer} & SegFormer~\cite{xie2021segformer}  & \shortstack{ADE20K, Cityscapes,\\ PASCAL VOC 2012}  & 2021-10-28\\
            \midrule
            MIT-B4~\cite{xie2021segformer} & SegFormer~\cite{xie2021segformer}  & \shortstack{ADE20K, Cityscapes,\\ PASCAL VOC 2012}  & 2021-10-28\\
            \midrule
            MIT-B5~\cite{xie2021segformer} & SegFormer~\cite{xie2021segformer}  & \shortstack{ADE20K, Cityscapes,\\ PASCAL VOC 2012}  & 2021-10-28\\
            \midrule
            UNet Convolutions & UNet~\cite{unet}  & Cityscapes  & 2015-05-18\\
            \midrule
            BEiT-Base~\cite{bao2021beit} & UPerNet~\cite{upernet}  & ADE20K  & 2022-09-03\\
            \midrule
            BEiT-Large~\cite{bao2021beit} & UPerNet~\cite{upernet}  & ADE20K  & 2022-09-03\\
            \midrule
            InternImage-Base~\cite{wang2023internimage} & UPerNet~\cite{upernet}  & \shortstack{ADE20K, Cityscapes,\\ PASCAL VOC 2012}  & 2023-04-17\\
            \midrule
            InternImage-Huge~\cite{wang2023internimage} & UPerNet~\cite{upernet}  & ADE20K  & 2023-04-17\\
            \midrule
            InternImage-Large~\cite{wang2023internimage} & UPerNet~\cite{upernet}  & ADE20K, Cityscapes  & 2023-04-17\\
            \midrule
            InternImage-Small~\cite{wang2023internimage} & UPerNet~\cite{upernet}  & \shortstack{ADE20K, Cityscapes,\\ PASCAL VOC 2012}  & 2023-04-17\\
            \midrule
            InternImage-Tiny~\cite{wang2023internimage} & UPerNet~\cite{upernet}  & \shortstack{ADE20K, Cityscapes,\\ PASCAL VOC 2012}  & 2023-04-17\\
            \midrule
            InternImage-XLarge~\cite{wang2023internimage} & UPerNet~\cite{upernet}  & ADE20K, Cityscapes  & 2023-04-17\\

        \bottomrule
    \end{tabular}
    }    
    \label{tab:model_zoo_dataset}
\end{table}
\Cref{tab:model_zoo_dataset} presents a comprehensive reference table for all semantic segmentation models used in our benchmarking. 
These methods include some of the first efforts in DL-based semantic segmentation methods like UNet~\cite{unet}, and some of the most recent SotA methods like InterImage~\cite{wang2023internimage}.
Each model is trained on the respective training subset of its dataset and evaluated on the corresponding validation set. The evaluations on 2D Common Corruptions are conducted using the validation sets.

\section{2D Common Corruptions}
\label{subsubsec:appendix:description:2dcc}
\cite{commoncorruptions} proposes introducing a distribution shift in the input samples by perturbing images with a total of 15 synthetic corruptions that could occur in the real world. 
These corruptions include weather phenomena such as fog, and frost, digital corruptions such as jpeg compression, pixelation, and different kinds of blurs like motion, and zoom blur, and noise corruptions such as Gaussian and shot noise amongst others corruption types.
Each of these corruptions can perturb the image at 5 different severity levels between 1 and 5.
The final performance of the model is the mean of the model's performance on all the corruptions, such that every corruption is used to perturb each image in the evaluation dataset.
Since these corruptions are applied to a 2D image, they are collectively termed 2D Common Corruptions.

\begin{figure}
    \centering
    \scalebox{0.55}{
    \begin{tabular}{cc}
     \rotatebox{90}{\textbf{\phantom{aaaaaaaa}Clean Input Image}}  & \includegraphics[width=0.75\linewidth, valign=b]{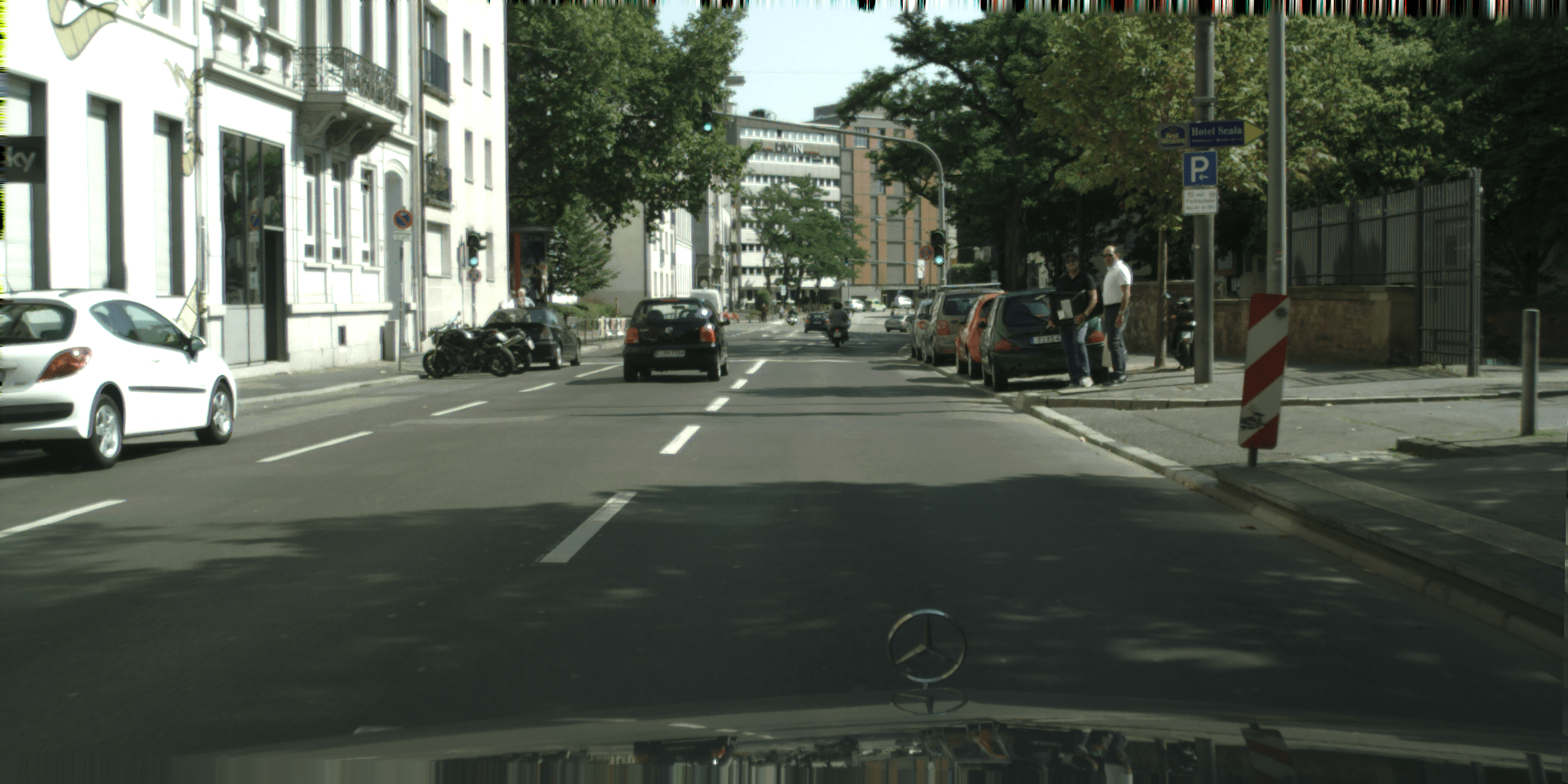} \\         
    \midrule
    \vspace{0.3em}
     \rotatebox{90}{\textbf{\phantom{aaaaaa}i.i.d.}}    &  \includegraphics[width=0.75\linewidth, valign=b]{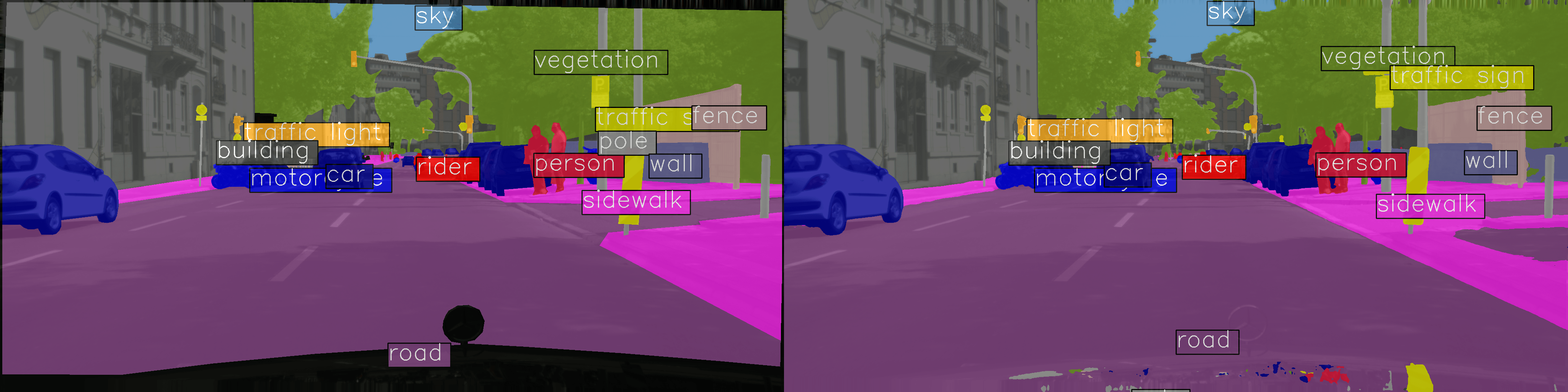} \\
    \vspace{0.3em}
     \rotatebox{90}{\phantom{aaaaaa}\textbf{Fog}}    &  \includegraphics[width=0.75\linewidth, valign=b]{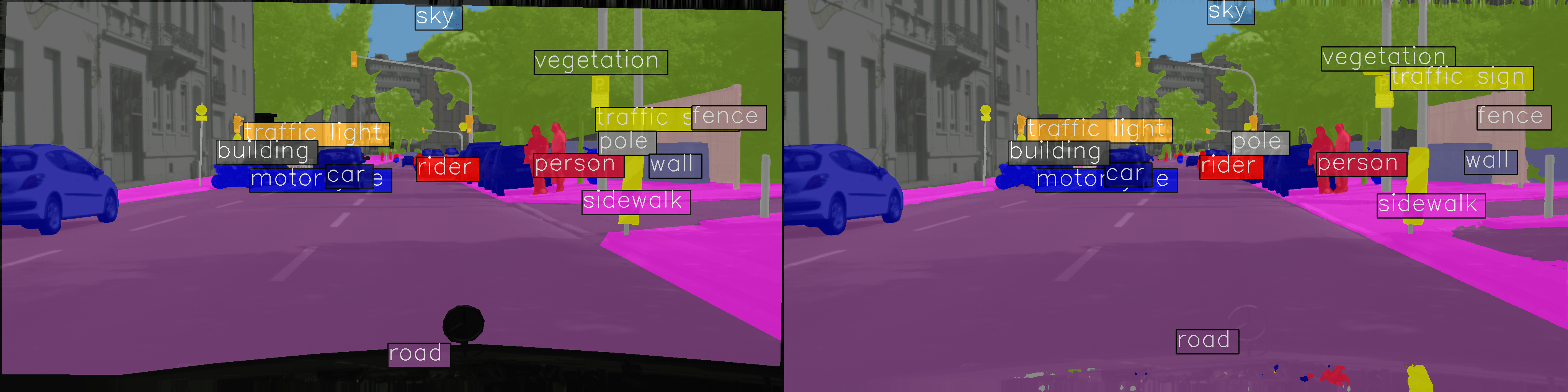} \\
    \vspace{0.3em}
     \rotatebox{90}{\phantom{aaaaaa}\textbf{Frost}}    &  \includegraphics[width=0.75\linewidth, valign=b]{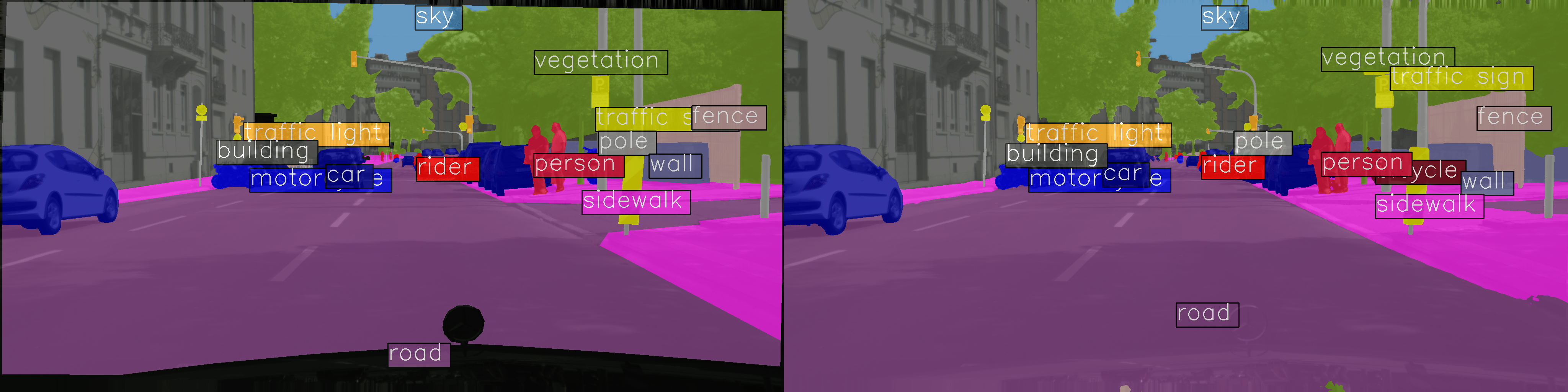} \\
    \vspace{0.3em}
     \rotatebox{90}{\phantom{aaa}\textbf{Motion Blur}}    &  \includegraphics[width=0.75\linewidth, valign=b]{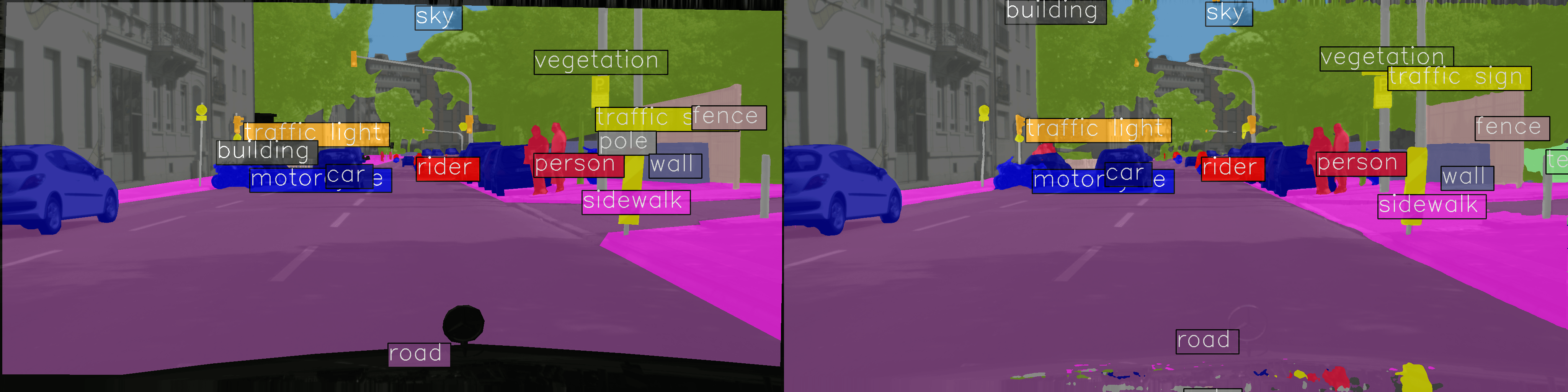} \\
     \vspace{0.3em}
     \rotatebox{90}{\phantom{aaa}\textbf{Zoom Blur}}    &  \includegraphics[width=0.75\linewidth, valign=b]{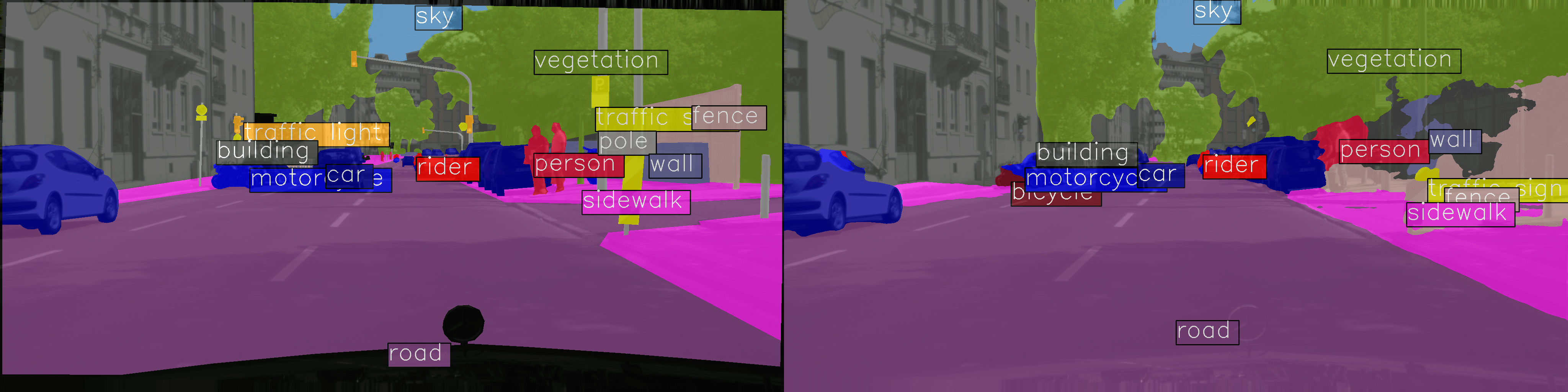} \\
    \vspace{0.3em}
     \rotatebox{90}{\phantom{aa}\textbf{Gaussian Noise}}    &  \includegraphics[width=0.75\linewidth, valign=b]{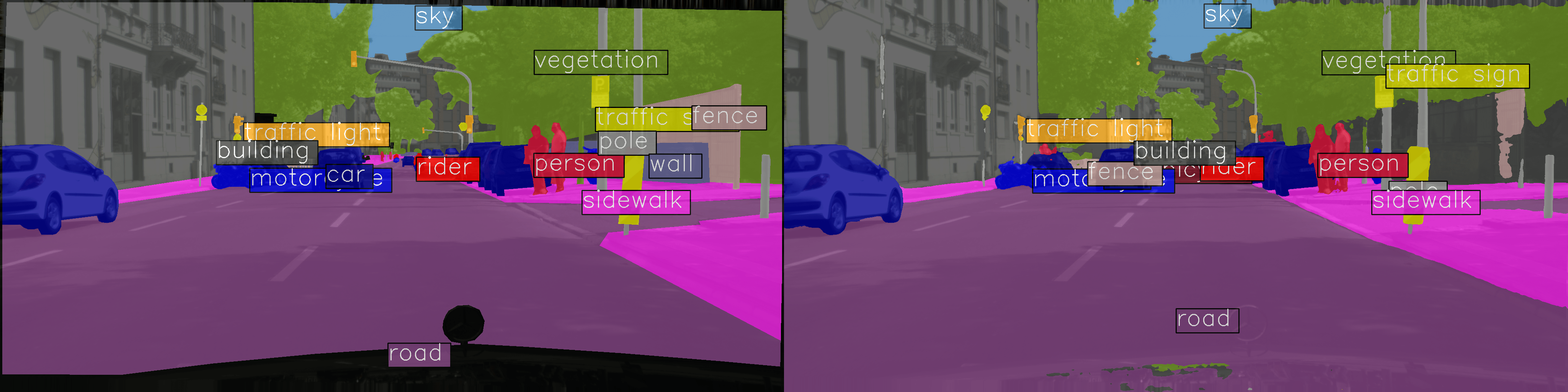} \\
    \vspace{0.3em}
     \rotatebox{90}{\phantom{aa}\textbf{Impulse Noise}}    &  \includegraphics[width=0.75\linewidth, valign=b]{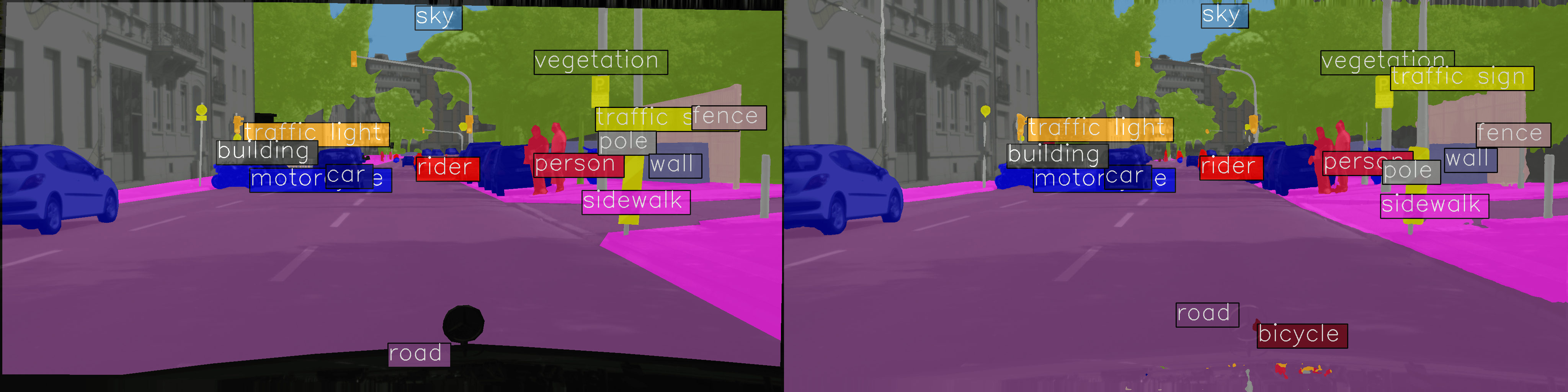} \\
    \vspace{0.3em}
     \rotatebox{90}{\phantom{aaaa}\textbf{Shot Noise}}    &  \includegraphics[width=0.75\linewidth, valign=b]{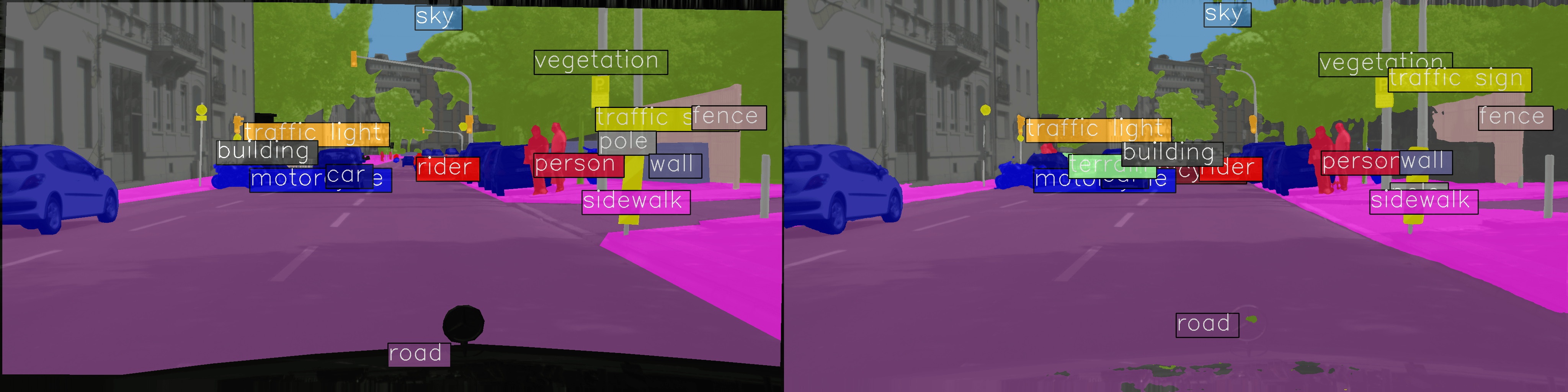} \\

    \end{tabular}
    }    
    \caption{Illustrating changes in prediction due to different 2D Common Corruptions on a randomly chosen input image from the \textbf{Cityscapes dataset}, when attaching the semantic segmentation method \textbf{InterImage-Base}. In the subfigures with semantic segmentation mask predictions, \textbf{Left: Ground Truth Mask}, and \textbf{Right: Predicted Mask}.}
    \label{fig:common_corruption_examples}
\end{figure}
We show examples of perturbed images over some corruptions and the changed predictions in \Cref{fig:common_corruption_examples}.

\begin{figure}
    \centering
   \centering
    \scalebox{1.0}{
    \begin{tabular}{@{}c@{\hspace{1.5mm}}c@{\hspace{1.5mm}}c@{}cc}
    & &
    Input Image & \hfil \phantom{aaaaaaaaaaa} Ground Truth & \hfil Prediction \\
    \multirow{2}{*}[3em]{\rotatebox{90}{Real World Corruption}} &
    \rotatebox{90}{\phantom{aaaaa} Night} &
    \includegraphics[width=0.3\linewidth]{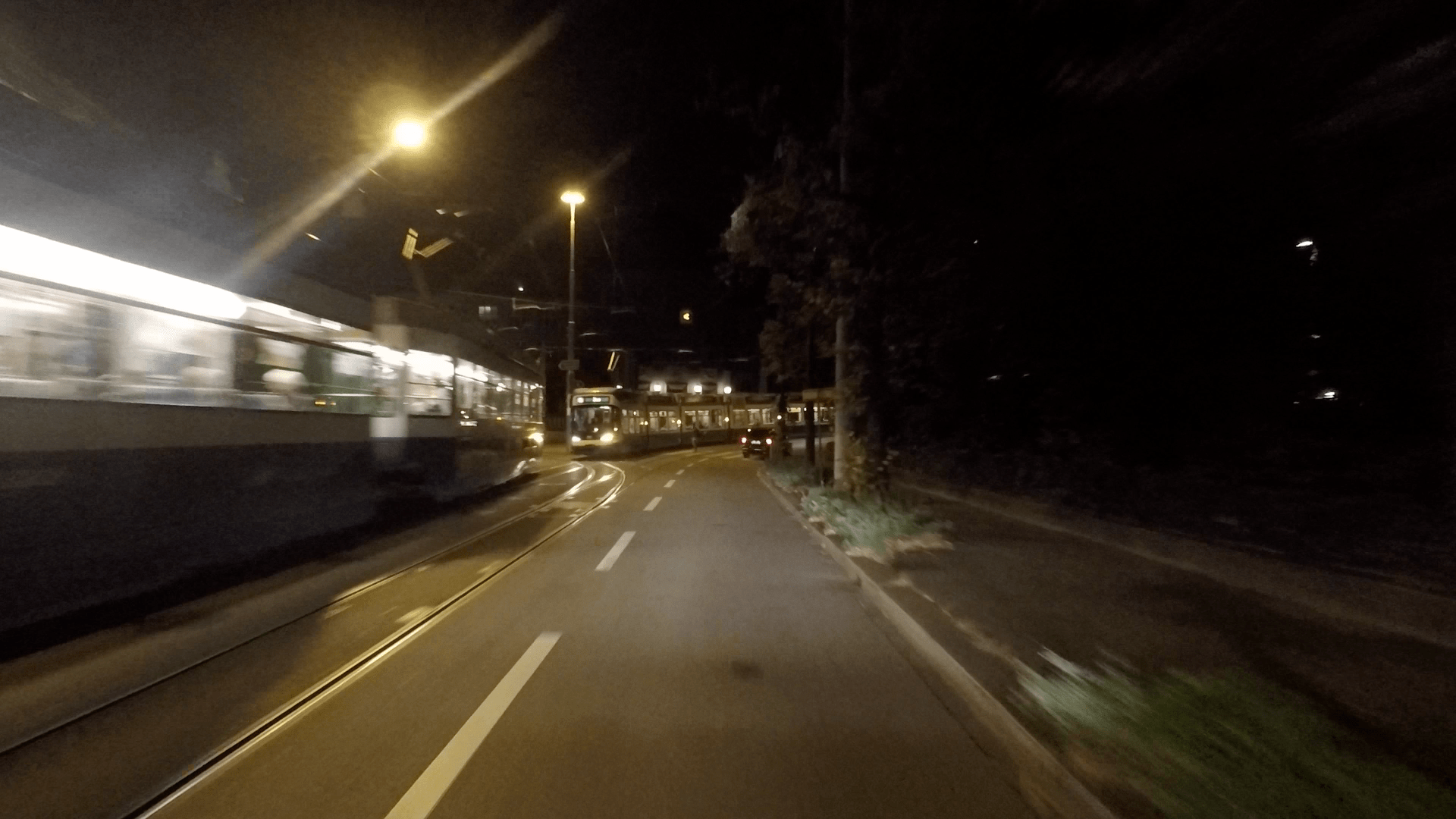}     
         & 
    \multicolumn{2}{c}{\includegraphics[width=0.6\linewidth]{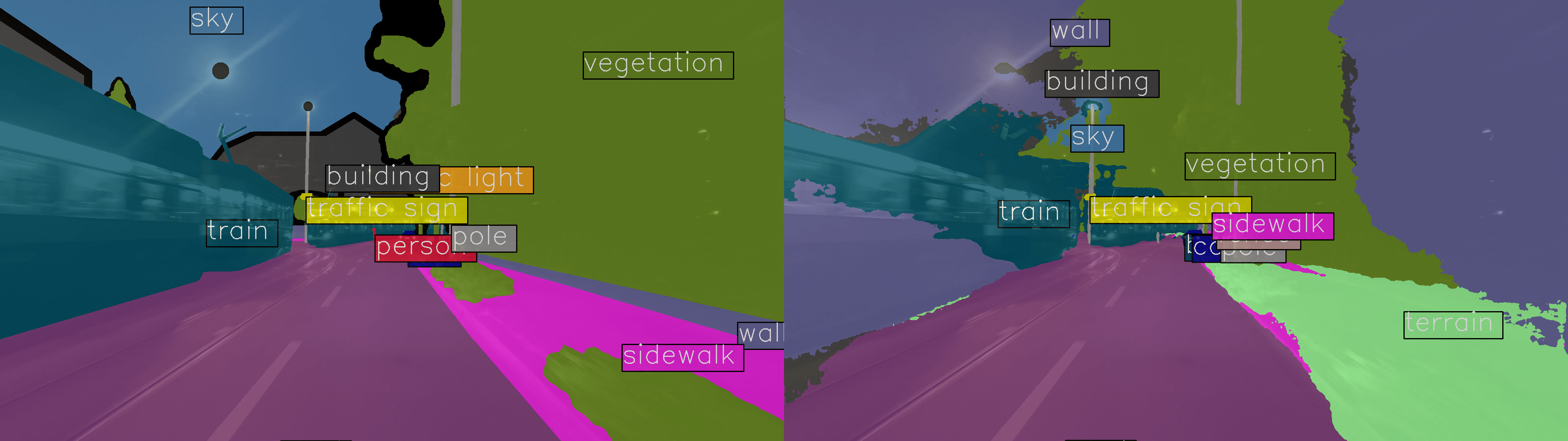}}
         \\

        &
        \rotatebox{90}{\phantom{aaaaaaa} Rain} &
    \includegraphics[width=0.3\linewidth]{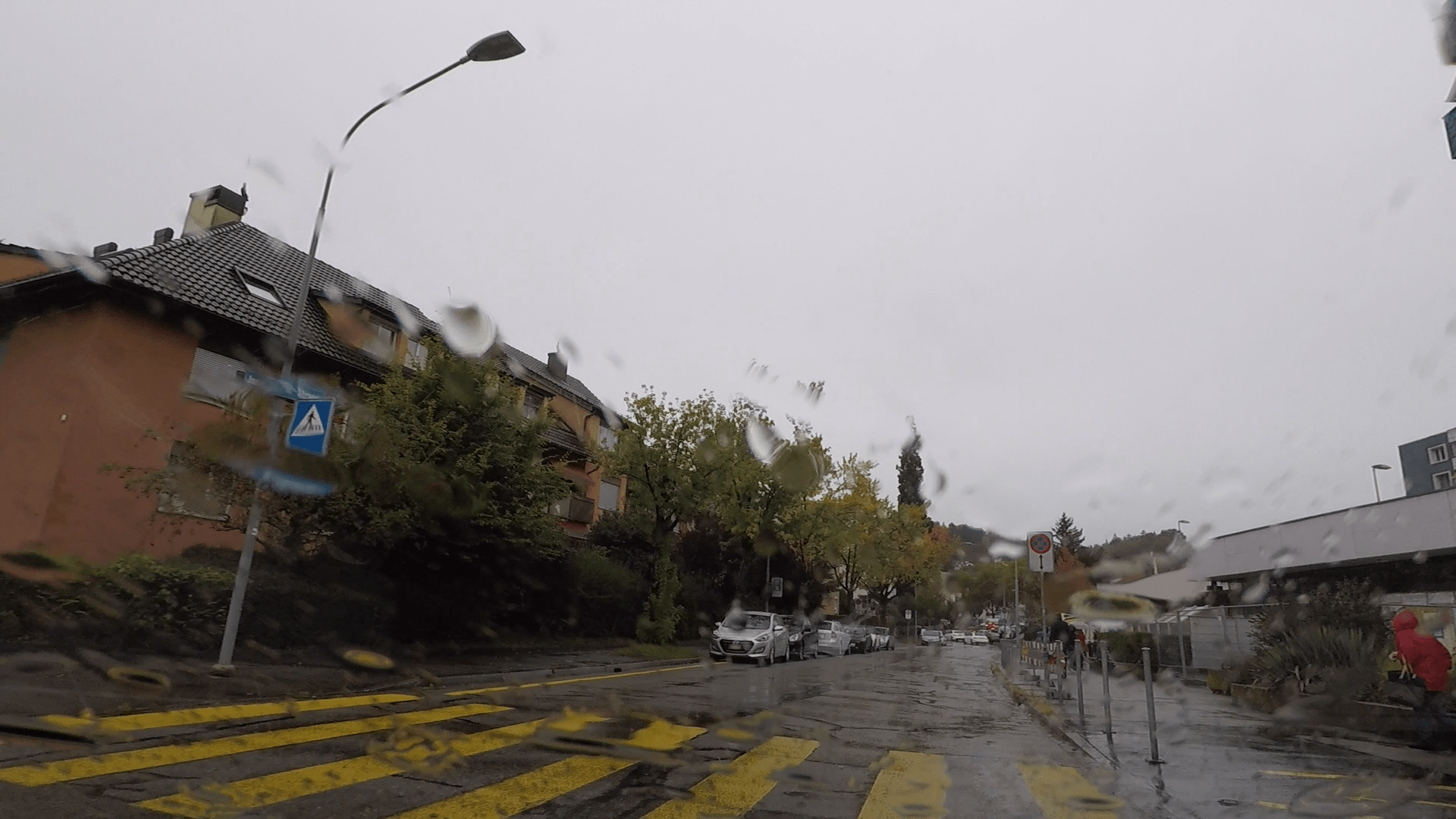}     
         & 
    \multicolumn{2}{c}{\includegraphics[width=0.6\linewidth]{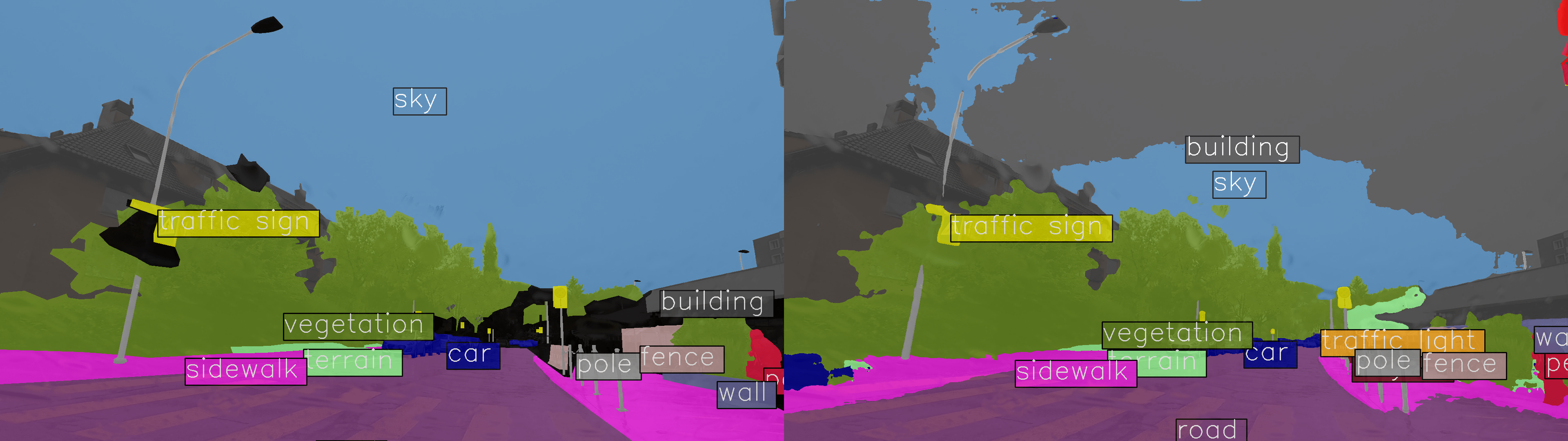}}
         \\


    \multirow{2}{*}[3em]{\rotatebox{90}{Synthetic Corruption}} &
    \rotatebox{90}{\phantom{aaa} Brightness} &
    \includegraphics[width=0.3\linewidth]{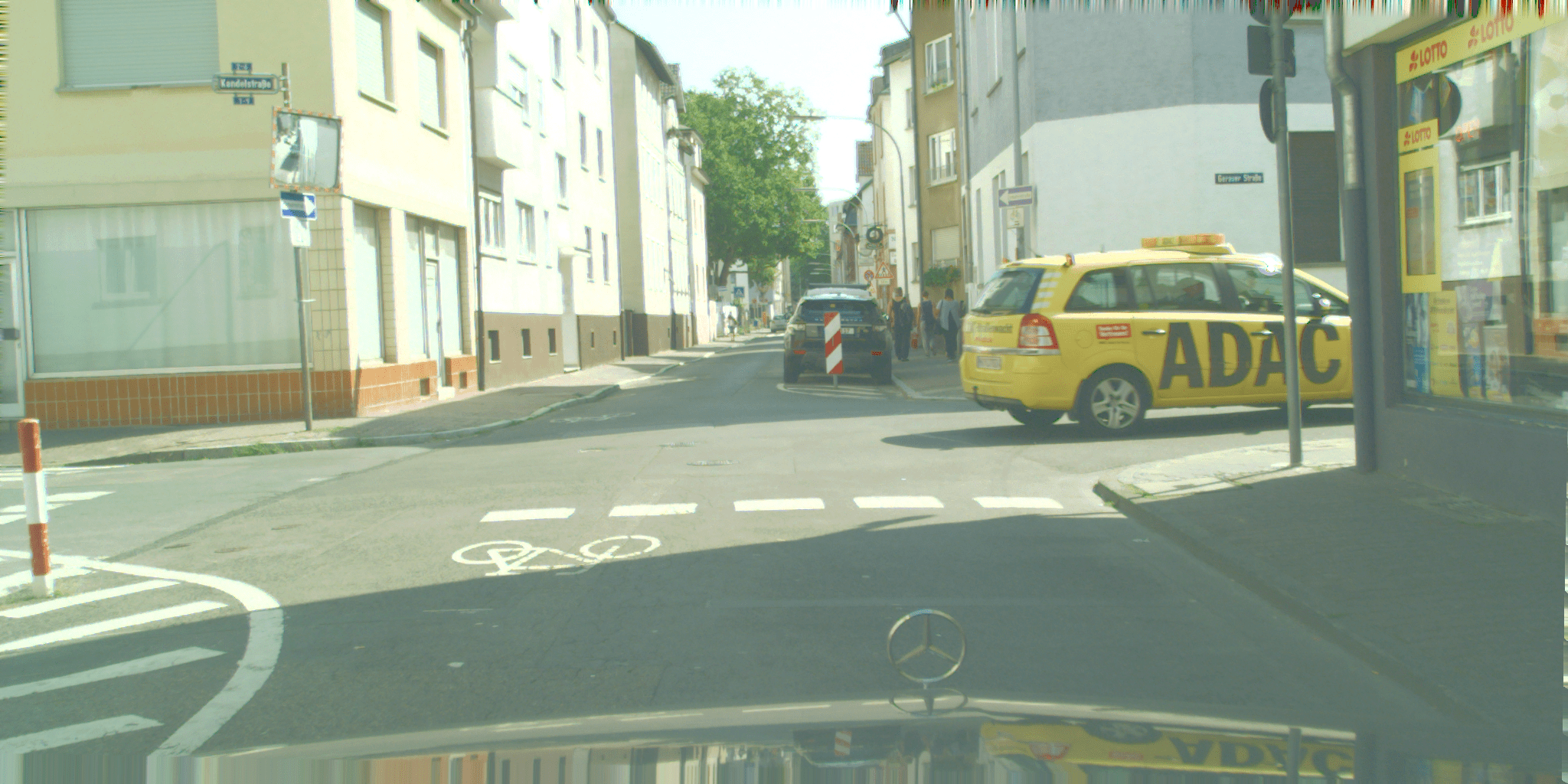}     
         & 
    \multicolumn{2}{c}{\includegraphics[width=0.6\linewidth]{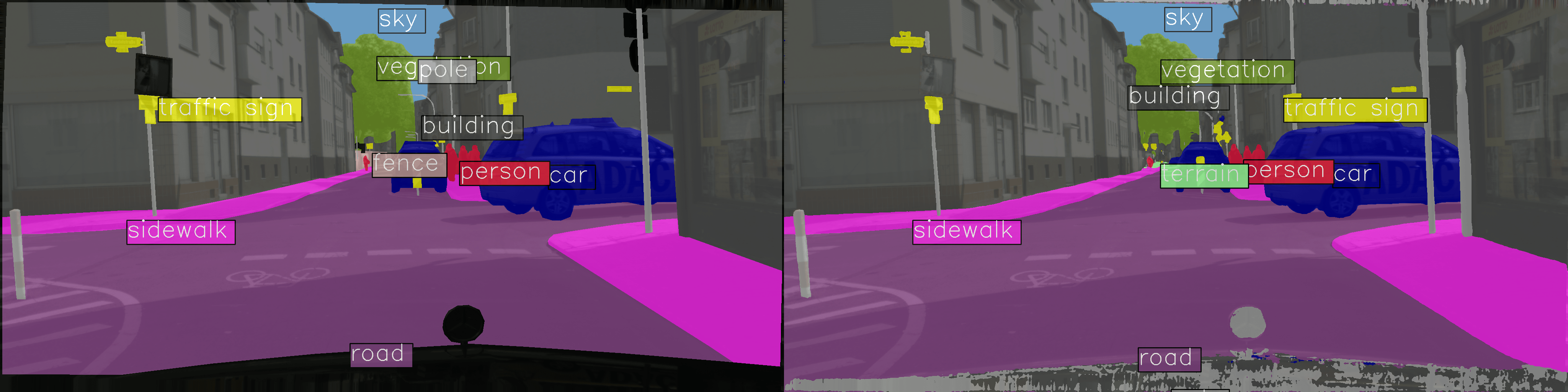}}
         \\

        &
        \rotatebox{90}{\phantom{aaaaaa} Frost} &
    \includegraphics[width=0.3\linewidth]{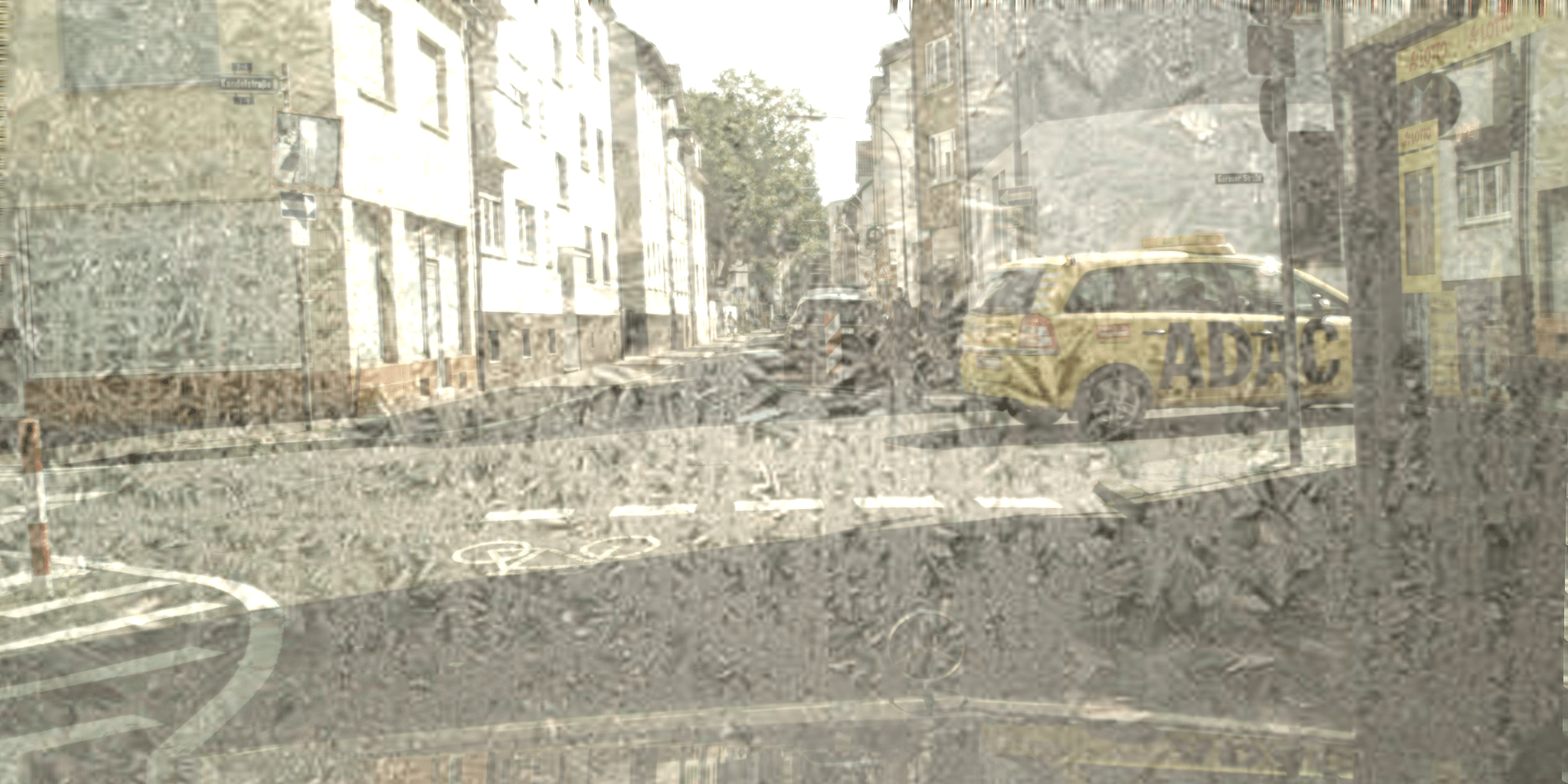}     
         & 
    \multicolumn{2}{c}{\includegraphics[width=0.6\linewidth]{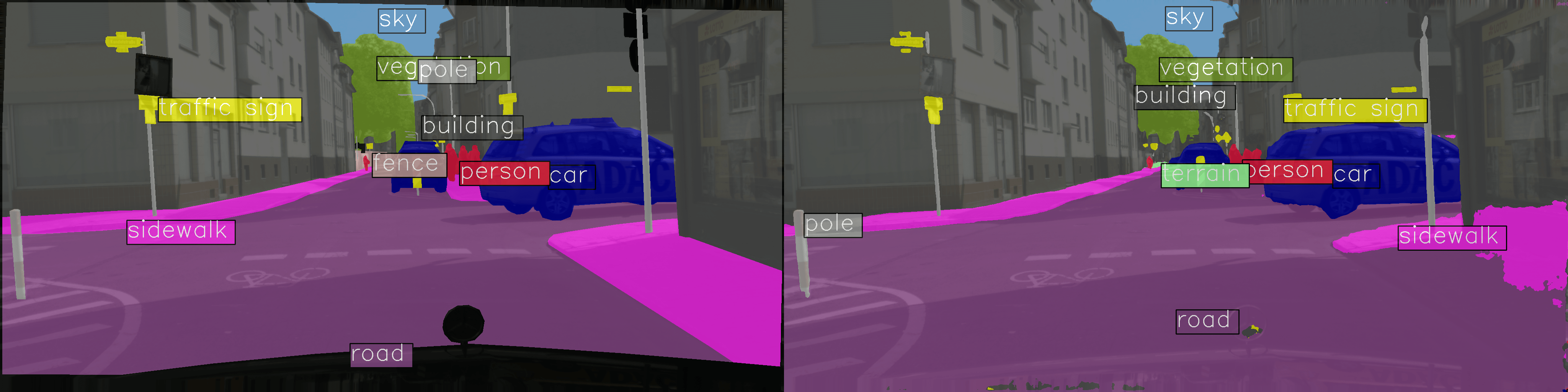}}
         \\
    \end{tabular}
    
    }
    \caption{An extension to \Cref{fig:teaser}, comparing images with weather corruptions captured in the wild (ACDC~\cite{acdc}) and images corrupted using synthetic corruptions~\cite{commoncorruptions} and the predictions using a Mask2Former~\cite{cheng2021mask2former} with a Swin-Base~\cite{liu2021Swin} backbone trained on the Cityscapes~\cite{cordts2016cityscapes} dataset.}
    \label{fig:teaser_extension}
\end{figure}
In \Cref{fig:teaser_extension}, we extend the visualizations from \Cref{fig:teaser}, additionally showing Night and Rain for ACDC, and Brightness and Frost for 2D Common Corruptions.

\extrafloats{100}

\section{Benchmarking Results}
\label{sec:appendix:additional_results}

\begin{figure}[ht]
    \centering
    \includegraphics[width=1.0\linewidth]{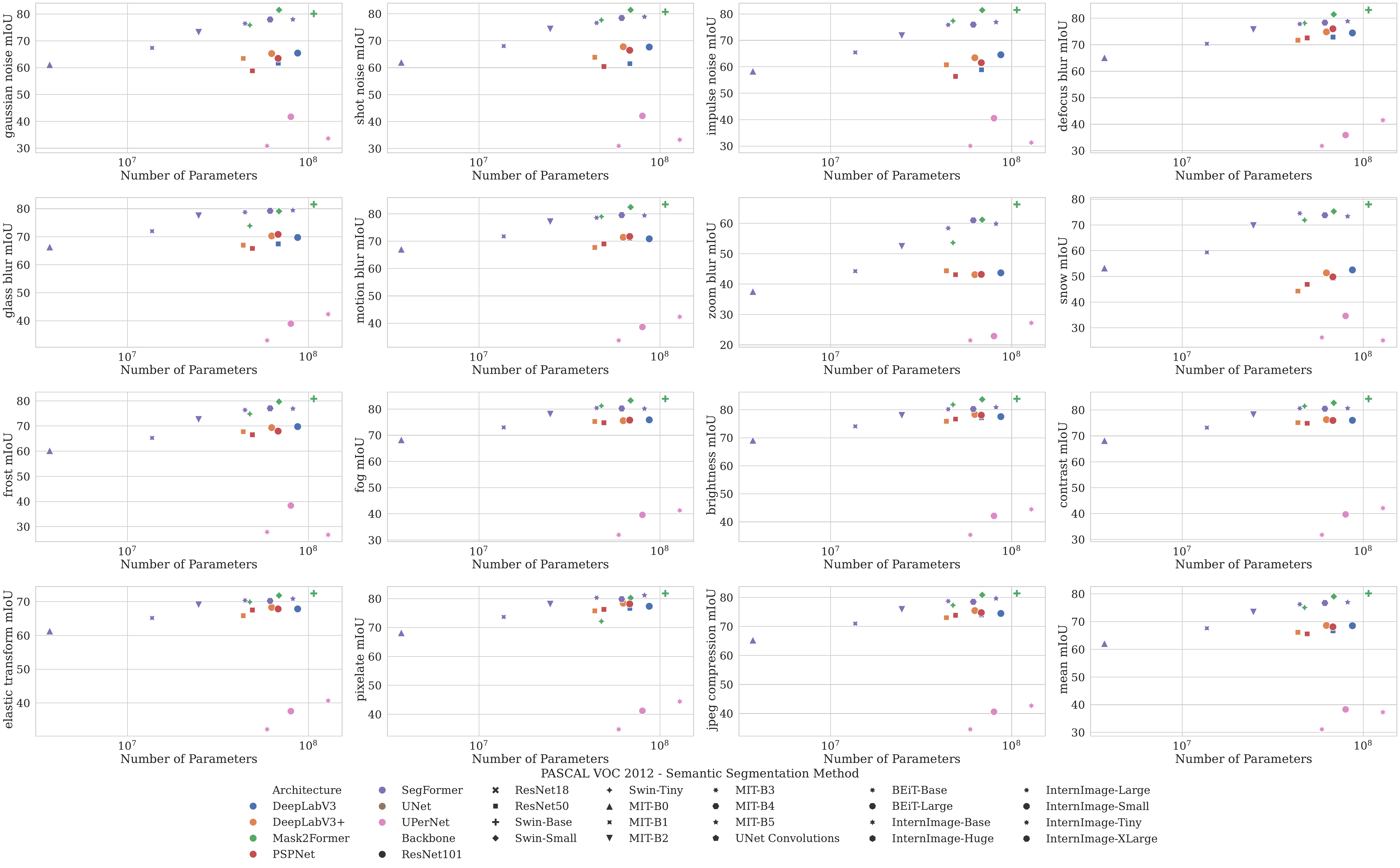}
    \caption{\textbf{Dataset used: PASCAL VOC2012}. The correlation in the performance of semantic segmentation methods against different 2D Common Corruptions. The respective axis shows the name of the common corruption used. Colors are used to show different architectures and marker styles are used to show different backbones used by the semantic segmentation methods. For the limited PASCAL VOC2012 evaluations we observe some correlation between the number of learnable parameters and the performance against common corruptions, however, more evaluations (more publicly available checkpoints) are required for a meaningful analysis.}
    \label{fig:2dcc_all_pascal_voc}
\end{figure}
\begin{figure}[ht]
    \centering
    \includegraphics[width=1.0\linewidth]{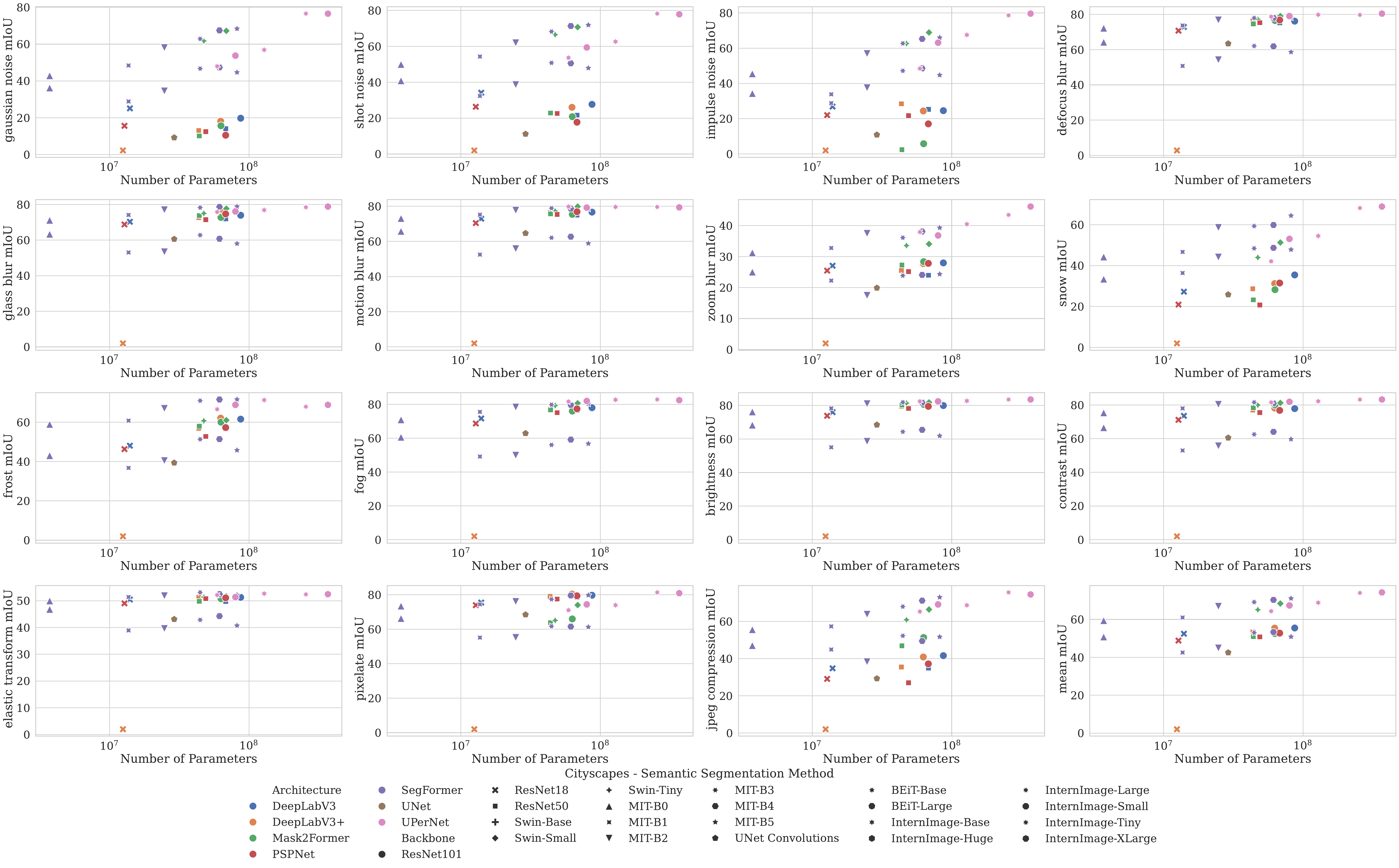}
    \caption{\textbf{Dataset used: Cityscapes}. The correlation in the performance of semantic segmentation methods against different 2D Common Corruptions. The respective axis shows the name of the common corruption used. Colors are used to show different architectures and marker styles are used to show different backbones used by the semantic segmentation methods. Except for DeepLabV3+ with a ResNet18 backbone, most other methods show a weak positive correlation between the number of learnable parameters used by a method and its performance against most of the common corruption. Multiple occurrences of an Architecture and Backbone pair are due to their evaluations being performed at two different crop sizes i.e.~512$\times$512, and 512$\times$1024.}
    \label{fig:2dcc_all_cityscapes}
\end{figure}

\begin{figure}[ht]
    \centering
    \includegraphics[width=1.0\linewidth]{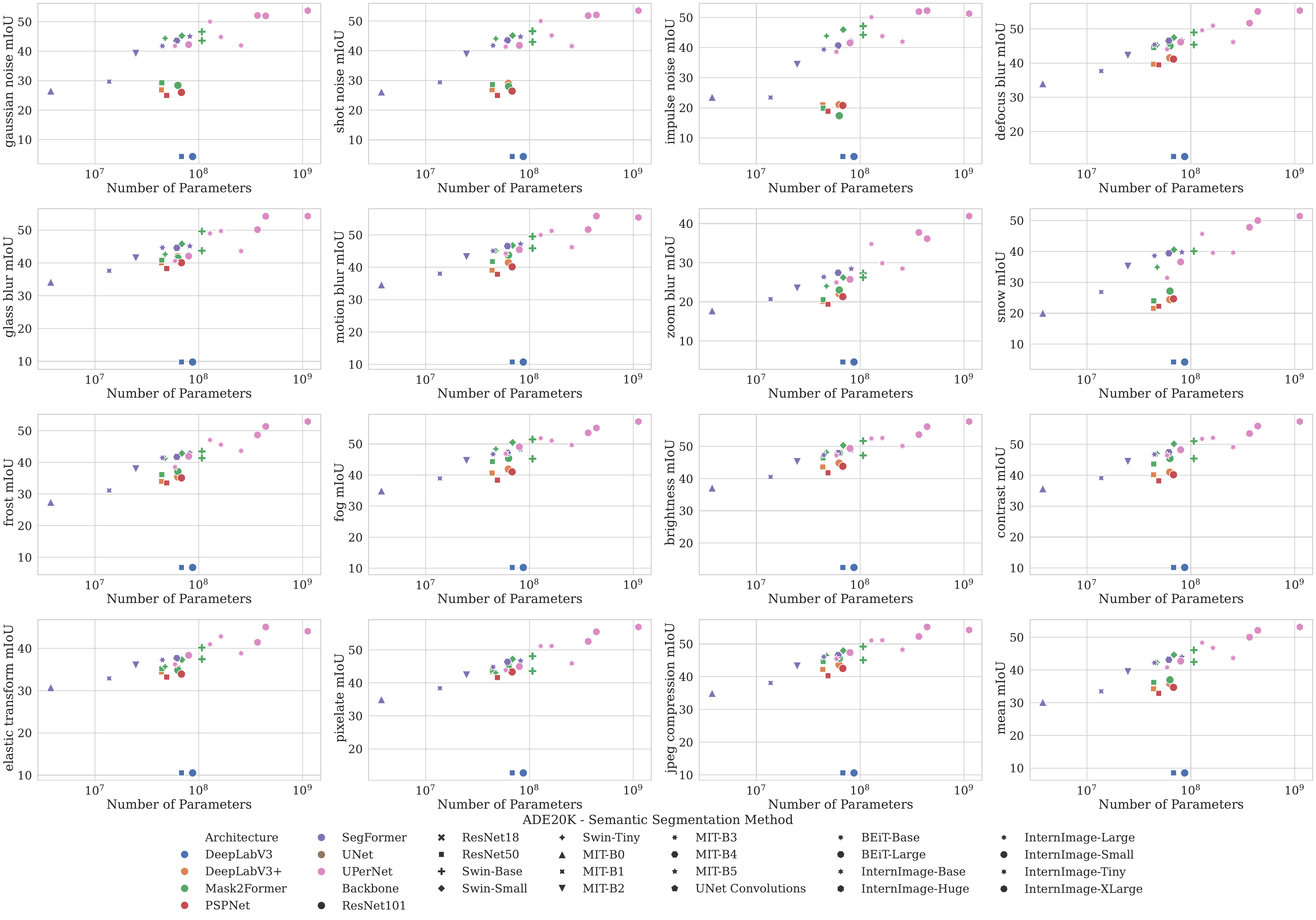}
    \caption{\textbf{Dataset used: ADE20K}. The correlation in the performance of semantic segmentation methods against different 2D Common Corruptions. The respective axis shows the name of the common corruption used. Colors are used to show different architectures and marker styles are used to show different backbones used by the semantic segmentation methods. Except for DeepLabV3, all other methods show some positive correlation between the number of learnable parameters used by a method and its performance against any common corruption.}
    \label{fig:2dcc_all_ade20k}
\end{figure}

Following, we include the results from the 2D Common Corruptions evaluations of all the semantic segmentation methods over all of the common corruptions, for PASCAL VOC2012 in \Cref{fig:2dcc_all_pascal_voc}, for Cityscapes in \Cref{fig:2dcc_all_cityscapes}, and for ADE20K in \Cref{fig:2dcc_all_ade20k}.

\section{Extension To The Related Work}
\label{sec:related_work_extension}
\citet{kamann2020benchmarking_semseg_ood} provide an OOD robustness benchmark for semantic segmentation.
While they use multiple backbone architectures, such as variants of ResNet~\cite{resnet}, MobileNet~\cite{mobilenet}, and Xception~\cite{chollet2017xception}, their evaluations are limited to the DeepLabV3+~\cite{deeplabv3+} architecture.
Our evaluated benchmark extends to multiple architectures and backbones, including recently proposed SotA methods like Mask2Former~\cite{cheng2021mask2former} and InternImage~\cite{wang2023internimage}.

\section{Future Work}
\label{subsec:conclusion:future_work}
Distribution shifts in the real world can be caused by multiple factors, one such factor is lens aberrations. \cite{muller2023classification} presents many such lens aberrations.
Additionally, \citet{3dcommoncorruptions} recently proposed 3D Common Corruptions that take scene depth into account to make corruptions more realistic-looking. We intend to extend our analysis to include these, enabling a more comprehensive robustness study.
Another valuable addition would be benchmarking semantic segmentation methods against adversarial attacks such as \cite{schmalfuss2022advsnow,scheurer2023detection,schmalfuss2022perturbationconstrained}.
Lastly, more in-depth analysis of the semantic segmentation methods, for example, as done by \cite{gavrikov2024training} for image classification methods, would help understand the models and their workings, especially in terms of their robustness performance.

\subsection{Limitations}
\label{subsec:conclusion:limitations}
Benchmarking the robustness of semantic segmentation methods is a computationally and labor-intensive endeavor. 
Thus, best utilizing available resources, we benchmark a limited number of settings. While more evaluations like correlation with different severity levels would be interesting, this is the most comprehensive robustness benchmark to date and instills interest to further improve our synthetic corruptions.


\end{document}